\PassOptionsToPackage{table, dvipsnames, xcdraw}{xcolor}
\documentclass[nonatbib]{elsarticle}

\journal{Experts Systems with Applications}

\usepackage[margin=2.5cm]{geometry}

\usepackage[utf8]{inputenc}

\usepackage{microtype}

\usepackage{etoolbox}
\robustify{\bfseries}

\makeatletter
\patchcmd{\ps@pprintTitle}
{submitted to}
{to appear in}
{}{}
\makeatother

\usepackage{pdflscape}

\usepackage{colortbl}

\usepackage{siunitx}
\DeclareSIUnit{\million}{M}
\DeclareSIUnit{\thousand}{K}
\usepackage{multirow}
\usepackage{booktabs}
\usepackage{datatool}

\newlength{\colsep}

\usepackage{csvsimple}
\usepackage{intcalc}

\usepackage{enumitem}

\usepackage{graphicx}
\makeatletter
\let\@originalcaption\@makecaption
\makeatother
\usepackage[font=footnotesize, labelformat=simple]{subcaption}
\usepackage{pgfplotstable}
\usepackage{pgfplots}
\usepackage{ifthen}
\usepackage{tikz}

\usetikzlibrary{
  shapes,
  decorations.pathmorphing,
  fit,
  backgrounds,
  calc,
  arrows,
  shadows,
  positioning,
  chains,
  scopes,
  matrix,
  shadings,
  shapes.arrows,
  shadows,
  arrows.meta,
  positioning,
  calc,
  external,
}

\usepgfplotslibrary{
  colorbrewer,
  statistics,
  groupplots,
}

\pgfplotsset{
  compat = 1.16,
  ymajorgrids,
  grid style={dashed}, %
  minimal plot grid/.style={
    y axis line style={opacity=0},
    axis x line*=bottom,
    x axis line style={black},
  },
  minimal plot grid,
  label style = {font=\footnotesize},
  tick label style = {font=\footnotesize},
  /pgf/declare function={
    Floor(\x) = round(\x-0.49);
  },
}

\newcommand\storelabel[2]{\expandafter\xdef\csname label#1\endcsname{#2}}
\newcommand\getlabel[1]{\csname label#1\endcsname}
\newcommand\gauss[4]{#4*(1/(#2*sqrt(2*pi))*exp(-((#3-#1)^2)/(2*#2^2)))}

\makeatletter
\let\c@author\relax
\makeatother

\usepackage[
backend=biber,
style=apa,
citestyle=authoryear,
maxcitenames=2,%
mincitenames=1,%
uniquename=init,
giveninits=true,%
sortcites,
sorting=nyt,
natbib=true,
eprint=false,
doi=false,
url=false,
isbn=false,
]{biblatex}
\AtBeginBibliography{\small}

\DefineBibliographyStrings{english}{%
  andothers = et al\adddot,
}

\DeclareSourcemap{
  \maps[datatype=bibtex, overwrite]{
    \map{
      \step[fieldsource=abstract]
      \step[fieldset=abstract, null]
    }
}}

\let\cite\citep

\DeclareNameFormat{labelname:poss}{%
  \nameparts{#1}%
  \ifcase\value{uniquename}%
  \usebibmacro{name:family}{\namepartfamily}{\namepartgiven}{\namepartprefix}{\namepartsuffix}%
  \or
  \ifuseprefix
  {\usebibmacro{name:first-last}{\namepartfamily}{\namepartgiveni}{\namepartprefix}{\namepartsuffixi}}
  {\usebibmacro{name:first-last}{\namepartfamily}{\namepartgiveni}{\namepartprefixi}{\namepartsuffixi}}%
  \or
  \usebibmacro{name:first-last}{\namepartfamily}{\namepartgiven}{\namepartprefix}{\namepartsuffix}%
  \fi
  \usebibmacro{name:andothers}%
  \ifnumequal{\value{listcount}}{\value{liststop}}{'s}{}}
\DeclareFieldFormat{shorthand:poss}{%
  \ifnameundef{labelname}{#1's}{#1}}
\DeclareFieldFormat{citetitle:poss}{\mkbibemph{#1}'s}
\DeclareFieldFormat{label:poss}{#1's}
\newrobustcmd*{\posscitealias}{%
  \AtNextCite{%
    \DeclareNameAlias{labelname}{labelname:poss}%
    \DeclareFieldAlias{shorthand}{shorthand:poss}%
    \DeclareFieldAlias{citetitle}{citetitle:poss}%
    \DeclareFieldAlias{label}{label:poss}}}
\newrobustcmd*{\citepos}{%
  \posscitealias%
  \textcite}
\newrobustcmd*{\Citepos}{\bibsentence\posscite}
\newrobustcmd*{\citespos}{%
  \posscitealias%
  \textcites}

\usepackage{xspace}
\makeatletter
\DeclareRobustCommand\onedot{\futurelet\@let@token\@onedot}
\def\@onedot{\ifx\@let@token.\else.\null\fi\xspace}

\def\eg{e.g\onedot} \def\Eg{E.g\onedot}
\def\ie{i.e\onedot} \def\Ie{I.e\onedot}
\def\cf{cf\onedot} 
\def\etc{etc\onedot} \def\vs{vs\onedot}
\def\wrt{w.r.t\onedot} 
\def\etal{et al\onedot} \def\aka{a.k.a\onedot}
\makeatother

\usepackage{fontawesome5}

\usepackage{scalerel}
\definecolor{orcidlogocol}{HTML}{A6CE39}
\usetikzlibrary{svg.path}
\tikzset{
  orcidlogo/.pic={
    \fill[orcidlogocol] svg{M256,128c0,70.7-57.3,128-128,128C57.3,256,0,198.7,0,128C0,57.3,57.3,0,128,0C198.7,0,256,57.3,256,128z};
    \fill[white] svg{M86.3,186.2H70.9V79.1h15.4v48.4V186.2z}
    svg{M108.9,79.1h41.6c39.6,0,57,28.3,57,53.6c0,27.5-21.5,53.6-56.8,53.6h-41.8V79.1z M124.3,172.4h24.5c34.9,0,42.9-26.5,42.9-39.7c0-21.5-13.7-39.7-43.7-39.7h-23.7V172.4z}
    svg{M88.7,56.8c0,5.5-4.5,10.1-10.1,10.1c-5.6,0-10.1-4.6-10.1-10.1c0-5.6,4.5-10.1,10.1-10.1C84.2,46.7,88.7,51.3,88.7,56.8z};
  }
}
\DeclareRobustCommand\orcidlogo[1]{\href{https://orcid.org/#1}{\ensuremath{\mbox{\scalerel*{%
\begin{tikzpicture}[yscale=-1, transform shape]%
\pic{orcidlogo};%
\end{tikzpicture}%
}{1}}}}}

\usepackage{url}
\usepackage[bookmarks=false]{hyperref}

\hypersetup{
  linkcolor  = red!50!black,
  citecolor  = blue!80!black,
  urlcolor   = violet!85!black,
  colorlinks = true,
}

\begin{document}

\begin{frontmatter}

\title{On the Pitfalls of Learning with Limited Data: A Facial Expression Recognition Case Study}
\tnotetext[code]{Code available at \protect\url{https://gitlab.com/mipl/learning-with-limited-data}.}

\author[unicamp]{Miguel Rodr\'iguez~Santander\corref{equal}}%
\ead{miguel.rodriguezs@ieee.org}
\author[unicamp]{Juan Hern\'andez~Albarrac\'in\corref{equal}}%
\ead{juan.albarracin@ic.unicamp.br}
\cortext[equal]{Equal contribution}
\author[unicamp]{Ad\'in Ram\'irez~Rivera\corref{cor1}}%
\cortext[cor1]{Corresponding author}
\ead{adin@ic.unicamp}

\address[unicamp]{Institute of Computing, University of Campinas, Campinas, SP, Brazil}

\begin{abstract}
Deep learning models need large amounts of data for training.
In video recognition and classification, significant advances were achieved with the introduction of new large databases.
However, the creation of large-databases for training is infeasible in several scenarios.
Thus, existing or small collected databases are typically joined and amplified to train these models.
Nevertheless, training neural networks on limited data is not straightforward and comes with a set of problems.
In this paper, we explore the effects of stacking databases, model initialization, and data amplification techniques when training with limited data on deep learning models' performance.
We focused on the problem of Facial Expression Recognition from videos.
We performed an extensive study with four databases at a different complexity and nine deep-learning architectures for video classification.
We found that (i)~complex training sets translate better to more stable test sets when trained with transfer learning and synthetically generated data, but their performance yields a high variance; (ii)~training with more detailed data translates to more stable performance on novel scenarios (albeit with lower performance); (iii)~merging heterogeneous data is not a straightforward improvement, as the type of augmentation and initialization is crucial; (iv)~classical data augmentation cannot fill the holes created by joining largely separated datasets; and (v)~inductive biases help to bridge the gap when paired with synthetic data, but this data is not enough when working with standard initialization techniques.
\end{abstract}


\end{frontmatter}

\section{Introduction}
\label{sec:introduction}
The development of computational intelligence techniques grew in recent years due to the broad adoption of deep learning models and increasing data availability.
The amount of data used to train deep learning models is crucial to avoid model under- or over-fitting.
Nevertheless, several domains (or problems) do not have large amounts of data available due to acquisition costs.
Hence, in these limited scenarios, deep learning solutions are incipient or, if they already exist, their results are limited due to the lack of labeled data.

Although standard techniques, such as dropout~\cite{Srivastava2014} or batch normalization~\cite{Szegedy2016}, are widely used to overcome the well-known over- and under-fitting problems, the silver bullet for every generalization problem is increasing the quantity of training data.
However, when no big data is available from a single source, there are several ways to increase data.
On the one hand, one can agglomerate several small datasets, \ie, where data from similar databases is compiled.
Another option is to do transfer learning, cross-domain adaptation, or cross-database training. Data on similar data-rich domains are used to create robust models that are subsequently fine-tuned on the limited-data problem.
And there is data augmentation, where new training examples are derived from the limited dataset through an augmentation function (commonly, another neural network is used).

It is important to point out that some of the work that explore limited-data setups are mainly surveys.
Thus, their models' quantitative comparison is of compiling nature~\cite{Lu2020, Shorten2019, Wang2020a}.
Although other works perform dedicated experiments to evaluate the impact of their techniques, the authors' focus is on one particular augmentation/initialization technique to improve the performance, and not on their combinations nor their contribution to the final result.
Only two works conducted an empirical study in which they combine more than one technique:
\citet{Chen2019} model complexity and cross-domain generalization on recognized few-shot learning models, and \citet{Brigato2020} study the effects of data augmentation and model complexity.
Hence, there is no focus on understanding the impact of limited data in the performance of models.

To the best of our knowledge, the overwhelming majority of studies on the impact of techniques facing small data focus on improving their process to increase performance.
However, there is no systematic exploration of the effects of the data in the pipeline.
Thus, the generalization effects of applying data augmentation techniques or transfer learning to limited-data setups are not well understood.
Therefore, our work is the first to study transfer learning impacts, dataset-stacking, data augmentation, and their combinations for a visual task.

In this work, we present the shortcomings of training with limited data and analyze the effects of different inductive biases related to data augmentation and model initialization techniques.
We focus on the Facial Expression Recognition (FER) problem in videos, as it is a domain that lacks data in the majority of the available datasets and has a variety of setups (\eg, posed \vs non-posed expressions, accessories, lighting, \etc) that make the datasets significantly heterogeneous among them.
We experiment across four FER datasets with different levels of complexity and nine deep-learning architectures for video recognition.
Our study includes comparisons on the effectiveness between classical \vs semantic data augmentation, transfer learning \vs random initialization, 2D- \vs 3D-CNN architectures, and low- \vs high-variance data used for dataset stacking.

Our main results are related to the characteristics of the used datasets.
For instance, the complexity of the training sets influences the learned model's performance, but they are correlated with the method used to augment the data from them.
On the one hand, complex training sets are better with transfer learning and synthetic data generators at the cost of higher variance in the results.
On the other hand, more straightforward datasets produce more stable results regardless of the augmentation used, but the performance is lower than the fine-tuned methods.
Another main result is that merging or stacking datasets to create a large database is not necessarily an improvement. 
This result follows from the gaps that appear in the training data modes that challenge the models' training.
To avoid these holes in the stacked data, one needs to take special care to initialize and augment the data (which translates to better final performance). 
Moreover, we observed that classical data augmentation techniques fell short, avoiding the previously mentioned problems. 
However, inductive biases help to bridge this gap when paired with synthetic data.

\section{Related Work}
\label{sec:related-work}

Previous work devoted to study the impact of common techniques to cope with small data is not scarce~\cite{Shin2016, Tajbakhsh2016, Soekhoe2016, Shijie2017, Chen2019, Pinetz2019, Altan2019, Brigato2020, Shorten2019, Guo2020, Lu2020, Wang2020, Wang2020a}, and focuses on understanding the impact of either Transfer Learning techniques~\cite{Shin2016, Tajbakhsh2016, Soekhoe2016, Chen2019, Lu2020}, data augmentation~\cite{Shijie2017, Pinetz2019, Shorten2019, Wang2020a, Brigato2020, Altan2020, Lu2020}, or cross-domain adaptation~\cite{Guo2020, Wang2020, Chen2019, Lu2020}.
Other works that proposed semantic data augmentation or transfer learning approaches punctually explore such an impact in a standard experimental setup: they compare their performance against baselines without the proposed data augmentation or transfer learning solution~\cite{Milicevic2018, Menon2019, HuynhThe2019, Wang2019, Bozorgtabar2019, Zhang2018, Jena2020, Bowles2018, Zhang2019c, Zhang2019d}.
Naturally, such works are not extensive because they usually do not explore many datasets or various architectures.
Nevertheless, these works' primary focus is to advance their respective tasks instead of understanding the overall effect of the techniques and the limited-data. 
Thus, there is a need to understand the effect that training with limited data has on the overall learning framework and the results that limited-data brings.

In this section, we present existing solutions to cope with limited-data while training deep neural networks.  
The effects of these solutions are the object of study in this paper, namely, data augmentation and transfer learning.
Then, we pose the problem of Facial Expression Recognition through the lens of limited data and present a brief state of the works that deal with the issue through such a lens.

\subsection{Data Augmentation}
Data augmentation consists in creating new training data through a set of transformations on the existing data.
This technique has become one of the most popular for training deep architectures.
In this work, we use image processing techniques and synthetic data generation for data augmentation.

\subsubsection{Classical Data Augmentation}
Classical Data Augmentation increases the training data using simple transformations.
For images, such transformations include flipping, rotation, cropping, noise injection, and changes in lighting, among others~\cite{Simonyan2015, Srivastava2015, He2016, Huang2017}.
This technique is usually implemented online; \ie, once a training batch of the original data is sampled; each example has some probability of being transformed through one of the operations above.
Nowadays, modern data augmentation methods automatically select existing techniques~\cite{Cubuk2019}, corrupt features as augmentation~\cite{Maaten2013, Wang2019}, or apply class identity preserving transformations~\cite{Jaderberg2016, Ratner2017}.

\subsubsection{Synthetic Data Augmentation}
\label{sec:synth-data}
Other works approached the lack-of-data problem using Generative Adversarial Networks~(GANs) to perform semantic data augmentation on the training data, reducing the models' over-fitting.
They applied it to a set of common problems, such as classification~\cite{Perez2017, FridAdar2018},  segmentation~\cite{Bowles2018}, detection~\cite{Han2019, Han2020}, among others.
Beyond these tasks, training GANs with limited data is challenging since generating realistic results with small datasets becomes impossible due to the over-fitting problems in the neural network.
To solve these problems, for instance, \citet{Zhao2020} used transfer learning techniques to train GANs in problems with limited data. 
On the other hand, \citet{Karras2020} proposed a novel technique to reduce the discriminator over-fitting through strategies that reduce noise when adding classic data augmentation processes.

The advances in Deep Generative Models, in particular GANs~\cite{Goodfellow2014, Gulrajani2017, Brock2019, Arjovsky2017}, have made it possible to synthesize data with increasing realism.
In the video synthesis domain, recent works in video reenactment have made it possible to generate videos whose objects (or characters) of interest mimic other videos~\cite{Siarohin2019, Chan2019, Zhao2018, Bansal2018}.
In particular, works on face reenactment~\cite{Wu2018, Zakharov2019, Nirkin2019, Aberman2019} have attained realistic results.

Although these methods have the potential for taking data augmentation for video to a higher semantic level, this idea has been widely explored mostly in images~\cite{Shorten2019, Wang2019}.
Recent works started augmenting data based on GANs for video classification, with dynamical images, \ie, only one frame representing the whole video~\cite{Zhang2019}.
For this setting, we chose Monkey-Net by \citet{Siarohin2019} as the model to synthesize novel videos, due to its simplicity \wrt other reenactment methods and its self-supervised learning mechanism, \ie, the architecture learns keypoints and optical flow while learning to reenact videos.

\subsection{Transfer Learning}

Under the analogy of human cognition's capacity to transfer knowledge from one domain to another, Transfer Learning involves techniques to train models on tasks where large amounts of data are available and use the learned parameters in tasks where data is scarce.
With the growth of deep learning, and the easiness of reusing models in different tasks, transferring techniques, (such as fine-tuning) have become the basis of modern Transfer Learning.
Training models with large data sets allows the first layers to learn to extract generic features of the data, while the last layers extract specific features of the task to be solved.
Fine-tuning was born from this idea, where already-trained models are used and retrained, either entirely or only a subset of its layers, in order to adapt to the new task model.

In this work, we use the fine-tuning technique by re-training all layers of the models that were already pre-trained with massive databases---\eg, ImageNet~\cite{Deng2009}, UCF-101~\cite{Soomro2012}, Kinects-700~\cite{Carreira2017}.
This technique serves as an inductive bias on the models since it assumes that the parameters learned for large-dataset image recognition are useful for the FER problem in videos from small datasets.

\subsection{Facial Expression Recognition as a Limited Data Problem}

Facial expression recognition (FER) has been widely studied in, what we pose as, a limited data environment~\cite{Li2018, Huang2019}.
Existing approaches involve recognizing the facial expression through static images~\cite{Ryu2017, Lopes2017,  Kuo2018, Acharya2018, MarreroFernandez2019, RamirezRivera2013} or videos~\cite{Klaser2008, Zhao2007, Zhao2011, Liu2014, Liu2015, Jung2015, Kaya2017, Hasani2017, Hasani2017a, Zhang2017, Acharya2018, Yan2018,  Zhang2019, Zhang2019a, Kuo2018, Liang2019, RamirezRivera2015, RamirezRivera2015a, Liu2020}.
The latter is incredibly difficult due to lack of data, since existing databases~\cite{Kanade2000, Lucey2010, Aifanti2010, Dhall2011, Dhall2012, Pantic2005, Valstar2010, Zhao2011} have a limited amount of videos depicting the expressions.
While image-based methods have an advantage due to a larger amount of frames extracted from these databases, video-based methods struggle with a restricted set of data from which to learn the patterns.
On the other hand, methods that rely on handcrafted features~\cite{Kim2016, Kaya2017, Zhang2019} need fewer data to learn in contrast to fully neural-network-based end-to-end learning methods~\cite{Jung2015, Liu2015, Zhao2016, Mollahosseini2016, Zhang2017, Hasani2017, Hasani2017a, Ding2017, Kuo2018, Acharya2018, MarreroFernandez2019, Liang2019}.  
Thus, they tend to obtain better results in these limited data scenarios due to the inductive biases used on the descriptors.

The vast majority of the literature's solutions focus on performance metrics on the test set regardless of their generalization to other domains.
This metric chasing leads to methods that cannot be used in environments with characteristics not covered by the training database.
One of the methodologies used to demonstrate such generalization capacity is the cross-database validation that consists in carrying out a proposed model's training with one database and the validation on a different one.

The main problem with training deep learning models with small databases is that they tend to over-fit the training data and get poor results when performing tasks in real environments.
In the last few years, many large databases have been released for public use, which allowed the development of new models that are more efficient and with low over-fitting rates.
Models pre-trained with large databases are used to solve other literature problems that do not have large databases.
Unlike natural-image databases (many of which contain millions of examples), there is a significant amount of databases of facial expressions in video available for public use. 
Still, all of them have a limited amount of data.

\section{Experimental Design}
\label{sec:setup}

We are interested in understanding the impact of the standard techniques to aggregate and augment data for small datasets, which generally are not enough to train deep models without the risk of over-fitting.
One common approach is to create samples from the existing databases artificially (\ie, data augmentation).
Another approach is to include inductive biases to the model, in the form of pre-trained weights (\ie, fine-tunning or transfer learning).
Another approach consists of stacking databases from different setups to increase (\ie, augment) the data.
These approaches can be either applied independently or combined in the same training pipeline.

Fine-tuning is commonly used in several methods in the literature.
It is easy to start with existing models and further tune them for related problems.
However, there is no attention paid to this transfer's limits in the literature when performed on limited data.
Moreover, novel methods with enough computing power (and data) rely on random initialization to avoid introducing biases from prior problems or data.

Data augmentation has been less studied but widely used in practice when no sustained effort to produce large amounts of data exist.
Another practice is to pool several databases (or data capture setups) together.
However, this practice introduces heterogeneous biases and sources of error that are not leveraged by many samples that can be used reliably to train the model.

Our main objective is to analyze the effect of these two sources of error when training with limited data, \ie, how to initialize the model, and how to augment the data.
In particular, we selected the facial expression recognition problem as it has several databases with a limited amount of samples and different setups and variable challenges.
Thus, stacking these databases will introduce different levels of uncertainty and variability that the models may not handle.

We propose to contrast deep learning classifiers for video (based on 2D and 3D convolutions and with different recurrent mechanisms) by using two initialization methods, two data augmentation techniques, and stacking the databases in different ways (using a hold-out database and hold-out cross-validation folds to get other ideas of the generalization).
Then, we discuss the observed results and conclude them.

\subsection{Proposed Experiments}
\label{sec:setup:proposed}
We executed a series of experiments to compare different network setups' performance and their generalization when faced with limited data.

\textbf{Model Comparison.} The objective of this experiment is to obtain state-of-the-art models' training metrics when training with limited data.
We decided to perform an ablation between each model using different types of parameter initialization---Xavier~\cite{Glorot2010} and transfer learning---and different types of data augmentation---classical and GAN-based.
Each of these comparisons was made for each database used in the study.

\textbf{Model Generalization.} To obtain metrics on the generalization of our models, we decided to perform cross-database experiments.
These experiments serve to compare model performances and assess their generalization capability.
To measure the level of generalization, we proposed two cross-database experiments:
\begin{itemize}
	\item Classic cross-database experiment: we trained with a database and tested on the others separately.
	\item All $k$-fold cross-validation: we created a $k$-fold cross-validation experiment by mixing all the databases into one. 
			We carefully maintained the parallel folds from other $k$-fold cross-validation experiments to have paired experiments for comparison.
			With this test, we obtained insights on the generalization from the data sources when stacking them for learning versus when learning on constrained sets.
\end{itemize}

\subsection{Data Sets}
\label{sec:datasets}

For this study, we consider four datasets whose size does not reach the order of thousands:

\begin{itemize}
\item \textbf{Extended CK+}~\cite{Lucey2010}. It consist of \num{593} sequences with \num{123} subjects with seven emotion labels (anger, contempt, disgust, fear, happiness, sadness, and surprise) recorded on a controlled environment.

\item \textbf{MMI}~\cite{Pantic2005, Valstar2010}. It contains \num{205} sequences from \num{30} subjects with six labels.
This database is not recorded in the wild, but the subjects present variations of worn accessories and movements during the expression performance.

\item \textbf{MUG}~\cite{Aifanti2010}. It contains \num{897} RGB videos of \num{52} different subjects and six expression classes: anger, disgust, fear, happiness, sadness, and surprise.  All the subjects are performing all the expressions.

\item \textbf{OULU-CASIA}~\cite{Zhao2011}. It comprises six expressions on three illumination conditions with \num{480} sequences per setting (\num{80} subjects with six expressions each).
\end{itemize}

\subsection{Classical and Deep Learning Architectures}
\label{sec:architectures}

For this work, we decided to compare (exhaustively, to the best of our efforts) the generalization capabilities of existing (and mostly used) deep learning architectures by training them with limited data.
Since we are interested in video tasks, particularly for FER, we selected 2D convolutional networks paired with recurrent models, 3D convolutional networks that deal with the sequence as a whole, and 3D convolutional networks paired with recurrent models to handle the sequence.

To assess the difference of performance between 3D-CNN-based architectures and LSTM-based ones, we opted for well-known ``vanilla'' architectures, with repeatable yet straightforward data setups.
Our objective is to make comparable scenarios, unlike some state-of-the-art works that lack open code or have complex loss functions with incomplete information that difficult a homogeneous comparison due to different training setups.

\subsubsection{Two-Dimensional Convolutional LSTM-based Architectures}
The 2D convolutional architecture was designed for image classification and commonly trained on ImageNet~\cite{Deng2009}.
We decided to use VGG16~\cite{Simonyan2015}, Inception V3~\cite{Szegedy2015, Szegedy2016, Szegedy2016a}, and ResNet-18 and~-101~\cite{He2016} due to their widespread use.
To adapt these architectures to work with videos, it was necessary to add recurrence.
We opted to use an LSTM~\cite{Hochreiter1997} that receives the extracted feature vectors by the 2D convolutional network for each frame and then calculate the recurrence relationships between these vectors, obtaining a classification label for the video.

\subsubsection{Three-Dimensional Convolutional Architectures}
The 3D convolutional architectures were designed for video tasks, especially video classification.
We decided to use C3D~\cite{Tran2015}, I3D~\cite{Carreira2017}, and ResNet3D-18 and~-101~\cite{Hara2018} since they obtain excellent results on UCF-101~\cite{Soomro2012} or Kinetics~\cite{Carreira2017}.
These models work over the whole video, so there is no need to add a recurrence.

\subsubsection{Three-Dimensional Convolutional LSTM-based Architectures}
The main difference between the two types of architecture presented above is the use of recurrent cells.
Three-dimensional convolution-based architectures extract spatiotemporal features with fixed time windows, while 2D-convolution-LSTM-based architectures extract spatial and temporal features independently, without time windows size limitations.
With that in mind, we propose the use of smaller temporal windows for the 3D convolutions and further process the encodings with a recurrent cell.
For this, we propose to use the C3D~\cite{Tran2015} architecture with an LSTM~\cite{Hochreiter1997} cell at the end.
We call this architecture C3D-Block-LSTM\@.

\subsection{Parameters}
\label{sec:parameters}

\begin{table}[tb]
  \centering
  \sisetup{
    table-format = 1,
  }
  \caption{Hyperparameters of the commonly used architectures that we included in our study.}
  \label{tab:models:parameters}
  \scriptsize
  \setlength{\colsep}{4pt}
%  \resizebox{\linewidth}{!}{%
  \begin{tabular}{%
      @{ }l@{\hspace{\colsep}}%
      l@{\hspace{\colsep}}%
      S@{\hspace{\colsep}}%
      S@{\hspace{\colsep}}%
      S@{\hspace{\colsep}}%
      S@{\hspace{\colsep}}%
      S@{\hspace{\colsep}}%
      S@{ }%
    }
    \toprule
    \textbf{Model name} & \textbf{Type} & \textbf{Frame size} & \textbf{Batch size}  & \textbf{Layers} & \textbf{LSTM size} & \textbf{GPUs}  & \textbf{Parameters}\\
    \midrule
    VGG16-LSTM & Conv2D & {(224,224)} & 10 & 16 & 1024 & 1 & 155M  \\
    InceptionV3-LSTM & Conv2D & {(300,300)} & 3 & 47 & 512 & 2 & 33M  \\
    ResNet18-LSTM & Conv2D & {(224,224)} & 10 & 18 & 512 & 1 & 13M \\
    ResNet101-LSTM & Conv2D & {(224,224)} & 3 & 101 & 512 & 2 & 46M  \\
    C3D & Conv3D & {(100,100)} & 10 & 10 & {--} & 1 & 78M \\
    I3D & Conv3D & {(226,226)} & 5 & 47 & {--} &  2 & 12M  \\
    ResNet3D-18 & Conv3D  & {(100,100)} & 10 & 18 & {--} & 1 &  33M \\
    ResNet3D-101 & Conv3D & {(100,100)} & 10  & 101 &  {--} &  1 & 85M  \\
    C3D-Block-LSTM & Conv3D & {(100,100)} & 10 & 10 
    & 1024 & 1 &  66M  \\
    \bottomrule
  \end{tabular}%}
\end{table}

The hyperparameters used in each network are the same ones presented by the original authors.
In models based on two-dimensional convolutions, an LSTM layer was added, followed by a classifier with three dense layers, where the last layer output is equal to the same number of classes from the database used.
We show the model-dependent parameters in Table~\ref{tab:models:parameters} (for implementation details of each model see the Appendix~\ref{app:models}).
We trained them using Adam~\cite{Kingma2014} with a learning rate of \num{1e-5}, weight decay of \num{5e-3}, dropout~\cite{Srivastava2014} with probability of \num{0.8}, and cross-entropy as a loss function.
We used $5$-fold cross-validation to obtain each model's metrics by dividing each database into five different subgroups whose divisions were person-independent.
We trimmed the videos using the most representative $25$~frames of the expression except for the C3D and I3D architectures that used $16$ and $64$~frames, respectively (see Fig.~\ref{fig:database:comparison} for the most representative parts on each database).
We used PyTorch~1.0~\cite{Paszke2017} to code and train all the models.
The hardware used for the experiments were two NVIDIA Titan Xp graphic cards.

\begin{figure}[tb]
\centering
\resizebox{\linewidth}{!}{\input{imgs/emotion_evolution.tex}}
\caption{%
The evolution of facial expressions in videos commonly presents a Gaussian-bell-like behavior, in which each sequence starts with a neutral expression and then evolves (onset) to a determined facial expression (apex) and returns to neutral (offset).
The observed temporal evolution is different in each database.
\Eg, sequences in CK+~\cite{Kanade2000, Lucey2010} ends at the apex, while AFEW~\cite{Dhall2011, Dhall2012} contains clips showing different parts of the evolution, and MUG~\cite{Aifanti2010}, MMI~\cite{Pantic2005, Valstar2010} and OULU-CASIA~\cite{Zhao2011} show the full evolution.
}
\label{fig:database:comparison}
\end{figure}

\section{Model Comparison with Limited Data}
\label{sec:model-limited-data}

The experimental setup proposed for this study is based on three scenarios, which are: (i)~single-database (Section~\ref{sec:single-database}), (ii)~merged-database (Section~\ref{sec:merged-database}), and (iii)~cross-dataset (Section~\ref{sec:cross-database}).
For each scenario, we provide a four-dimensional analysis, in which each dimension corresponds to a family of variations on the training scheme.
These dimensions are:
\begin{enumerate}
  \item \textbf{Model Initialization.} As introduced above, we explore Random Initialization~(RI) through Xavier~\cite{Glorot2010}, and Fine-tunning~(FT).
  \item \textbf{Data Augmentation Scheme.} We explore Classical Data Augmentation~(DA), and Semantic Data Augmentation with Synthesized Data~(SD).
  \item \textbf{Models.} We use the nine models introduced above: VGG16-LSTM, InceptionV3-LSTM, ResNet18-LSTM, ResNet101-LSTM, C3D, C3D-Block-LSTM, I3D, ResNet3D-18, and ResNet3D-101.
  \item \textbf{Databases.} As described above, we use four datasets: CK+, MMI, MUG, and OULU\@.
\end{enumerate}

Due to the large volume of results obtained, we provide tables with detailed results in Appendixes~\ref{app:single-model}, \ref{app:merged-database}, and~\ref{app:cross-databases}, along with the results of exhaustive statistical tests in Appendix~\ref{app:stat-test}.
We compare the performances of different models setups in Figs.~\ref{fig:comparison-model-and-datasets-unified}, \ref{fig:comparison-model-unified}, \ref{fig:comparison}, and~\ref{fig:cross-database}.

Every configuration in the setup described above (\ie every initialization-augmentation-model-database combination) underwent a $5$-fold cross-validation scheme.
As suggested in the literature~\cite{Demvsar2006, Dietterich1998}, we performed Wilcoxon Signed-Rank tests~\cite{Demvsar2006} between every pair of experiments, at $5\%$ significance level.
We show the $p$-values of those tests in Appendix~\ref{app:stat-test}.

The lack of transparency regarding model configuration and training details from most of the literature methods prevents us to safely compare the performance of the models used in this study with the ones that conform to state of the art.
Aspects like the train/validation/test split of the datasets, the number of folds used for cross-validation, the spatiotemporal cropping to pre-process the sequences, and the number of frames used for fixed-length models, must be aligned along with all the models for a fair comparison and few works report these details.

For that reason, a comparison between our performance metrics and the ones reported in the literature would be merely speculative and should not be considered as a proper comparison.

\subsection{Single-Dataset $k$-Fold Cross-Validation}
\label{sec:single-database}

\begin{figure}[tb]
  \centering
  {\begin{tikzpicture}%
\begin{groupplot}[%
  group style = {
    group size=1 by 1,
    horizontal sep = 1.5em,
    vertical sep = 1.5em,
    x descriptions at=edge bottom,
    y descriptions at=edge left,
  },
  width=.85\linewidth,
  height=5cm,
  cycle list/Set3,
  cycle list={
    Set3-E, Set3-F
  },
  boxplot/draw direction=y,
  enlarge x limits=0.04,
  xmajorgrids,
  xticklabels = {RI, FT, RI+DA, FT+DA, RI+SD, FT+SD},
  ylabel style = {align=center},
  ylabel = {Accuracy (\%)},
  ymin = 0,
  ymax = 1,
  boxplot={
    draw position={1/3 + Floor(\plotnumofactualtype/2) + 1/3*mod(\plotnumofactualtype,2)},
    box extend=0.25,
  },
  every boxplot/.style={mark=*,every mark/.append style={mark size=1pt}},
  xtick={0,1,2,...,6},
  x tick label as interval,
  x tick label style={
    align=center
  },
  legend entries = {{Single}, {Merged}},
  legend to name={grouplegend},
  legend style={
    font=\scriptsize,
    legend columns=2,
    draw=none,
  },
]
\def\myplots{}
\xappto\myplots{\noexpand\nextgroupplot[]}
\pgfplotstableread[col sep=comma]{./imgs/auto_ablation_merge_models_and_datasets/metadata.csv}\metadb
\pgfplotstableforeachcolumnelement{config}\of\metadb\as\config{%
  \foreach \folder [count=\i] in {auto_ablation_merge_models_and_datasets,cross_kfold_ablation_merge_models_and_datasets}{%
    \xappto\myplots{%
      \noexpand\begingroup
      \noexpand\pgfplotstableread[col sep=comma]{./tablesv2/\folder/\config.csv}\noexpand\csvdata
      \noexpand\pgfplotstabletranspose\noexpand\csvdata{\noexpand\csvdata}
      \noexpand\addplot+[boxplot, fill, draw=black!50] table[y index=1] {\noexpand\csvdata};
      \noexpand\endgroup
    }
  }
}
\myplots%
\end{groupplot}%
\path (group c1r1.north east) -- node[above, yshift=0cm]{\pgfplotslegendfromname{grouplegend}} (group c1r1.north west);
\end{tikzpicture}}%
  \caption{
    Aggregated results of Fig.~\ref{fig:comparison} per training configuration.
    We compare training and evaluation on the same database~(single) and using a merged database~(merged).
    We used random~(RI) and transfer learning by fine-tuning previously trained models~(FT) for initialization.
    We used affine transformations used in classical data augmentation~(DA) and synthetic data generated with models from the training partitions~(SD) for augmentation.
  }
  \label{fig:comparison-model-and-datasets-unified}
\end{figure}
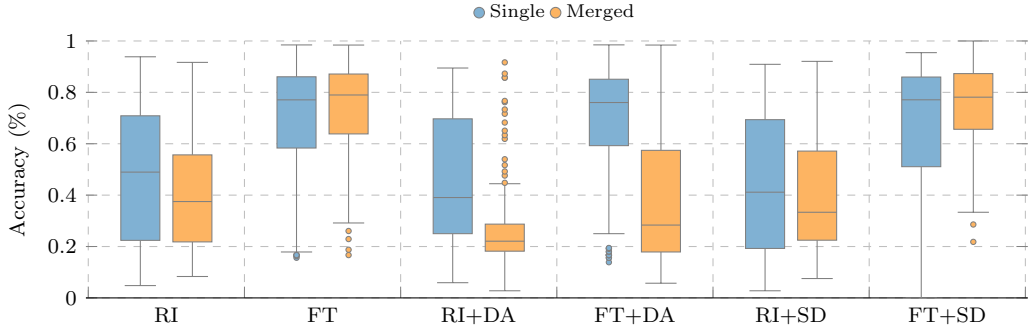

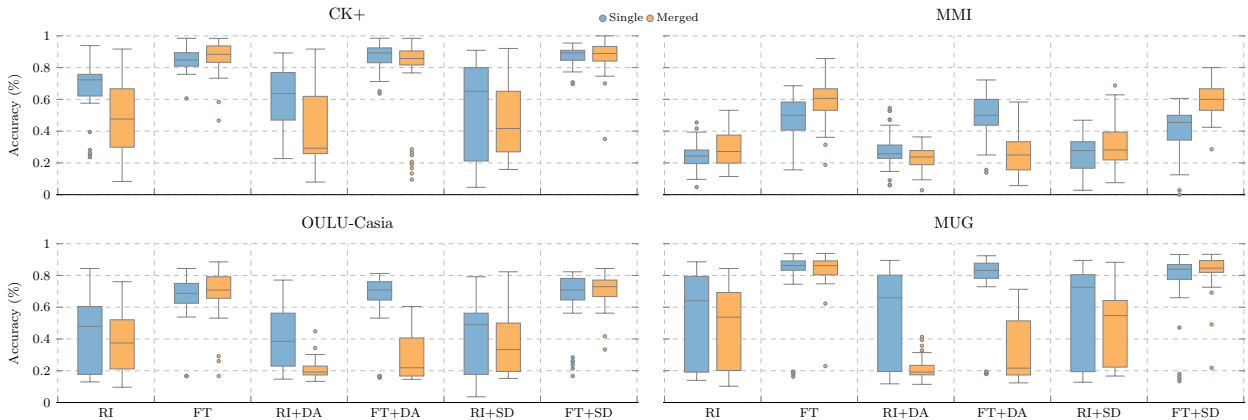
\begin{figure}[tb]
  \centering
  \resizebox{\linewidth}{!}{\begin{tikzpicture}%
\begin{groupplot}[%
  group style = {
    group size=2 by 2,
    horizontal sep = 1.5em,
    vertical sep = 3em,
    x descriptions at=edge bottom,
    y descriptions at=edge left,
  },
  width=.85\linewidth,
  height=5cm,
  cycle list/Set3,
  cycle list={
    Set3-E, Set3-F
  },
  boxplot/draw direction=y,
  enlarge x limits=.04,
  xmajorgrids,
  xticklabels = {RI, FT, RI+DA, FT+DA, RI+SD, FT+SD},
  ylabel style = {align=center},
  ylabel = {Accuracy (\%)},
  ymin = 0,
  ymax = 1,
  boxplot={
    draw position={1/3 + Floor(\plotnumofactualtype/2) + 1/3*mod(\plotnumofactualtype,2)},
    box extend=0.25,
  },
  every boxplot/.style={mark=*,every mark/.append style={mark size=1pt}},
  xtick={0,1,2,...,6},
  x tick label as interval,
  x tick label style={
    align=center
  },
  legend entries = {{Single}, {Merged}},
  legend to name={grouplegend},
  legend style={
    font=\scriptsize,
    legend columns=2,
    draw=none,
  },
]
\def\myplots{}
\foreach \dbname [count=\j]in {CK, MMI, OULU, MUG}{%
  \ifdefstring{\dbname}{CK}{\def\dbnameshow{CK+}}{\ifdefstring{\dbname}{OULU}{\def\dbnameshow{OULU-Casia}}{\def\dbnameshow{\dbname}}}
  \xappto\myplots{\noexpand\nextgroupplot[title={\dbnameshow}]}
  
  \pgfplotstableread[col sep=comma]{./imgs/auto_ablation_merge_models/metadata.csv}\metadb
  \pgfplotstableforeachcolumnelement{config}\of\metadb\as\config{%
    \foreach \folder [count=\i] in {auto_ablation_merge_models,cross_kfold_ablation_merge_models}{%
      \xappto\myplots{%
        \noexpand\begingroup
        \noexpand\pgfplotstableread[col sep=comma]{./tablesv2/\folder/\dbname/\config.csv}\noexpand\csvdata
        \noexpand\pgfplotstabletranspose\noexpand\csvdata{\noexpand\csvdata}
        \noexpand\addplot+[boxplot, fill, draw=black!50] table[y index=1] {\noexpand\csvdata};
        \noexpand\endgroup
      }
    }
  }
}\myplots%
\end{groupplot}%
\path (group c1r1.north east) -- node[above, yshift=0cm]{\pgfplotslegendfromname{grouplegend}} (group c2r1.north west);
\end{tikzpicture}}%
  \caption{
    Aggregated results of Fig.~\ref{fig:comparison} per dataset.
    We compare training and evaluation on the same database~(single) and using a merged database~(merged).
    We used random~(RI) and transfer learning by fine-tuning previously trained models~(FT) for initialization.
    We used affine transformations used in classical data augmentation~(DA) and synthetic data generated with models from the training partitions~(SD) for augmentation.
  }
  \label{fig:comparison-model-unified}
\end{figure}

For the single-database scenario, we replicated standard experiments in the literature performed within a single database.

Figures~\ref{fig:comparison-model-and-datasets-unified} and~\ref{fig:comparison-model-unified} show aggregated results to compare experimental setups (\ie, initialization plus augmentation) for all and each dataset, respectively.
The former aggregates paired experiments per method and database, while the latter aggregates only per method.
The results correspond to the mean and variance of the $5$-fold cross-validation setup.
We can see that all configurations with FT obtained significantly better results than any configurations with RI\@.
In general (see Fig.~\ref{fig:comparison-model-and-datasets-unified}), the augmentation methods (either DA or SD) seems to be irrelevant in RI (\cf Table~\ref{app:tab:st-merged-single-models unified-dataset}).
The only difference was observed in CK+, in which no augmentation presents a significant superiority over SD\@.
For FT, MUG showed a decrease in performance when using any data augmentation, MMI showed that DA and no augmentation achieved better results than SD, and CK+ only showed the superiority of SD over no augmentation.
OULU did not show any pattern.
Diving into more detail, the left column of Fig.~\ref{fig:comparison} shows the accuracy on each database by training the methods (shown with different colors) with several combinations of initialization-augmentation techniques (shown as groups within each database).

In summary, FT obtains better results than RI in all setups.
For RI, the impact of any data augmentation is virtually null (except for CK+).
On the other side, FT presents a more heterogeneous behavior \wrt the data augmentation technique and varies according to the dataset.

\subsubsection{Model Initialization}
As a first observation, the initialization method dramatically influences the results.
When using RI, the models obtain lower and less stable metrics (higher variance) when compared to FT, which shows lower variances, in general.
Compare the RI and FT groups in the left of Fig.~\ref{fig:comparison} for each method, or aggregated in Figs.~\ref{fig:comparison-model-and-datasets-unified} and~\ref{fig:comparison-model-unified}.
Additionally, the temporal changes are a challenge to these methods.
For instance, the CK+ and MUG databases are more stable because facial expressions are posed, unlike MMI and OULU\@.
Furthermore, this result is reflected in the generalized higher classification performance in the former databases.

When using FT, the CNN-based networks reuse their learned features and tune them to the small amounts of data.
It was evidenced that, for the 2D-CNN-based methods, FT yields significantly better performance than RI in all datasets, regardless of the type of data augmentation.
On the other hand, the 3D-CNN-based methods present a more complex behavior since the improvement of FT \wrt RI seems to depend on the data augmentation type, and the results differ between datasets.
For example, FT is significantly better than RI for CK+ and OULU when no augmentation is done and for DA, but this improvement is weaker for SD (\cf Fig.~\ref{fig:comparison-model-and-datasets-unified}).
FT improves in MMI when no data augmentation is used.
For DA and SD, this improvement is weaker.
Noticeably, the spread of variance is reduced and more consistent in MMI in contrast to CK+ or MUG\@.
The latter presents the same behavior as MMI, except for the I3D, ResNet3D-101 and ResNet3D-18 models (\cf Fig.~\ref{fig:comparison}).
I3D presented worst results for FT \wrt RI, and ResNet3D-101 and ResNet3D-18.
These three methods presented a weaker improvement of FT for all the datasets, but only in MUG that was evident.
Finally, note that by coupling the spatiotemporal features with a recurrence (\ie, a 3D block sequences processed through an LSTM on C3D-Block-LSTM), we obtained more unsatisfactory performance when compared with its purely 3D counterparts, and this difference is higher in the random initialization experiments.

When comparing 2D- versus 3D-based methods, we observed no significant difference in performance between the two families of methods.
Individually, ResNet18-LSTM and C3D attained the lowest performances, while the rest of the methods were similar.

Analyzing the different models' results for their depth, we observe that models that have deeper architectures (\ie, VGG16-LSTM, InceptionV3-LSTM, I3D, ResNet101-LSTM, and R3D-101) present a subtle superiority \wrt the shallow architectures (\ie, C3D, C3D-Block-LSTM, ResNet18-LSTM, and ResNet3D-18).
We observed that deep models took more advantage of fine-tunning than shallow ones, regardless of the data augmentation type, due to their bigger capacity, to which more data is beneficial.

In general, FT is more stable, \ie, with a smaller variance, in contrast to RI in the four datasets we tested (\cf Fig.~\ref{fig:comparison-model-and-datasets-unified}).
Moreover, FT obtains consistently better results than RI\@.

In summary, although FT presents a clear advantage \wrt RI, the latter presents a notably higher variance, and the significance of this improvement highly depends on the dataset, the model, and the augmentation technique.
The most stable datasets, namely CK+ and MUG, obtained the best classification accuracies.
For the 2D-CNN-based models, FT significantly outperforms RI in all datasets, regardless of the augmentation technique.
On the other side, for the 3D-based models, this improvement is more heterogeneous: when no augmentation is performed, the superiority of FT over RI is significant but weaker for DA and SD.
In general, there is no significant difference in the performance of 2D-CNN-based methods and 3D-CNN-based ones, while the LSTM+3D-CNN-method had a lower performance than the rest.
Finally, deep architectures present a slightly superior performance over shallow ones for FT, while this difference is less evident for RI.

\subsubsection{Data Augmentation}
\label{sec:data-aug}

\begin{figure}[tb]
  \centering
  \resizebox{\linewidth}{!}{\begin{tikzpicture}%
\begin{groupplot}[%
  group style = {
    group size=2 by 4,
    horizontal sep = .5em,
    vertical sep = 1em,
    x descriptions at=edge bottom,
    y descriptions at=edge left,
  },
  width=.85\linewidth,
  height=5cm,
  cycle list/Set3-9,
  boxplot/draw direction=y,
  enlarge x limits=.02,
  xmajorgrids,
  xticklabels = {RI, FT, RI+DA, FT+DA, RI+SD, FT+SD},
  ylabel style = {align=center},
  ymin = 0,
  ymax = 1,
  boxplot={
    draw position={1/10 + Floor(\plotnumofactualtype/9) + 1/10*mod(\plotnumofactualtype,9)},
    box extend=0.05,
  },
  every boxplot/.style={mark=*,every mark/.append style={mark size=1pt}},
  xtick={0,1,2,...,6},
  x tick label as interval,
  x tick label style={
    align=center
  },
  legend entries = {{VGG16-LSTM}, {InceptionV3-LSTM},{ResNet18-LSTM},{ResNet101-LSTM},{C3D},{C3D-Block-LSTM},{I3D},{ResNet3D-18},{ResNet3D-101}},
  legend to name={grouplegend},
  legend style={
    font=\scriptsize,
    legend columns=9,
    draw=none,
  },
]
\def\myplots{}
\foreach \dbname in {CK, MMI, OULU, MUG}{%
  \ifdefstring{\dbname}{CK}{\def\dbnameshow{CK+}}{\def\dbnameshow{\dbname}}
  \foreach \folder [count=\i] in {auto_ablation,cross_kfold_ablation}{%
    \ifnum \i=1
    \xappto\myplots{\noexpand\nextgroupplot[ylabel={\dbnameshow\noexpand\\Accuracy (\%)}]}
    \else
    \xappto\myplots{\noexpand\nextgroupplot[]}
    \fi
    \pgfplotstableread[col sep=comma]{./imgs/\folder/metadata.csv}\metadb
    \pgfplotstableforeachcolumnelement{config}\of\metadb\as\config{%
      \xappto\myplots{%
        \noexpand\begingroup
        \noexpand\pgfplotstableread[col sep=comma]{./tablesv2/\folder/\dbname/\config.csv}\noexpand\csvdata
        \noexpand\pgfplotstabletranspose\noexpand\csvdata{\noexpand\csvdata}   
      }
      \foreach \n in {1,...,9} {%
        \xappto\myplots{\noexpand\addplot+[boxplot, fill, draw=black!50] table[y index=\n] {\noexpand\csvdata};}
      }
      \xappto\myplots{\noexpand\endgroup}
    }
  }
}\myplots%
\end{groupplot}%
\path (group c1r1.north east) -- node[above, yshift=0cm]{\pgfplotslegendfromname{grouplegend}} (group c2r1.north west);
\end{tikzpicture}}%
  \caption{
    Accuracy of several models when trained with different combinations of initialization and augmentation techniques.
    We used random~(RI) and transfer learning by fine-tuning previously trained models~(FT) for initialization.
    We used affine transformations used in classical data augmentation~(DA) and synthetic data generated with models from the training partitions~(SD) for augmentation.
    We show training and evaluation on the same database~(left) and using a merged database~(right)---\cf Sections~\ref{sec:single-database} and~\ref{sec:merged-database}, respectively.
  }
  \label{fig:comparison}
\end{figure}
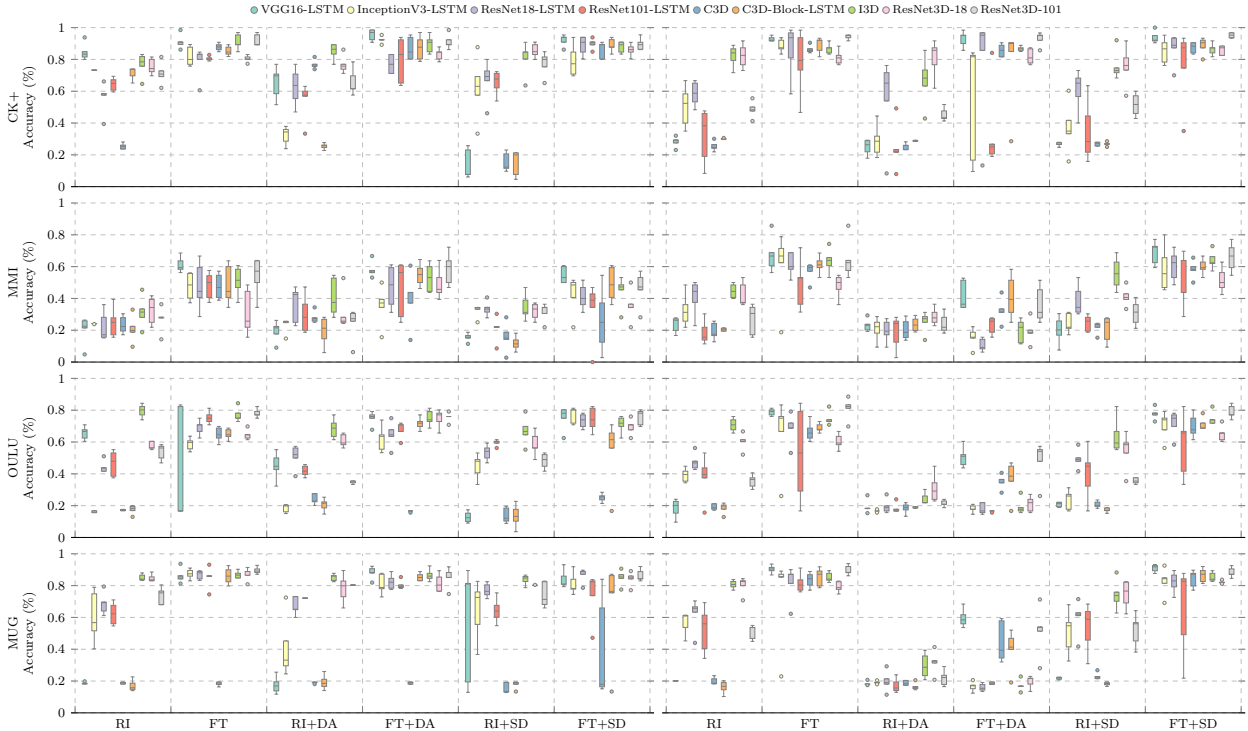

As described above, we considered two types of data augmentation: classical~(DA) and GAN-based~(SD).
For DA, we obtained the former with affine transformations on the images (\eg, rotations, and flip), commonly used in previous work.
For SD, we trained one generative model per experiment, namely Monkey-Net~\cite{Siarohin2019}, from the training partition, and then used it to generate novel examples as a source for data augmentation.
We selected these two augmentation methods due to their wide-spread use and limited ourselves to them due to computational resources to execute the experiments.
Table~\ref{app:tab:single-model} shows the detailed comparison between the impact of DA and SD on the model performance.

The impact of DA for RI depends on both the database and the model.
For example, for CK+, DA's effect in the C3D-block-LSTM, InceptionV3-LSTM, ResNet101-LSTM, ResNet3D-101, and VGG16-LSTM architectures is a decrease on accuracy performance that can be considered as significant, while the converse is true for C3D and I3D.
For ResNet18-LSTM, no significant difference was evidenced (\cf Fig.~\ref{fig:comparison}).
For OULU, DA increased accuracy on C3D and decreased it for I3D, ResNet3D-101, and VGG16-LSTM\@.
In general, for no database, RI presented any significant difference in performance when comparing RI+DA\@.
Nevertheless, we observed an increment in the variance \wrt the non-augmented version in all datasets (\cf Fig.~\ref{fig:comparison-model-and-datasets-unified}).

For FT, DA shows a non-significant general improvement (\cf Table~\ref{app:tab:st-single-models-unified}).
Note, however, that for MUG, DA significantly reduced the accuracy of FT\@.
There are more outliers in OULU (\cf Fig.~\ref{fig:comparison} FT \vs FT+DA), on which C3D-block-LSTM and ResNet3D-18 got their accuracy increased, while C3D and ResNet101-LSTM got it decreased.

Regarding the impact of SD for RI, CK+ presented an increase in accuracy only for ResNet3D-18 and decreased for C3D, C3D-block-LSTM, and VGG16-LSTM\@.
We observed no significant changes in the rest of the models.
Similarly, we observed no significant impact for the MUG dataset, except for a decrease in the performance of ResNet3D-18.
OULU presented increments in InceptionV3-LSTM, ResNet101-LSTM and ResNet18-LSTM, and decrements in I3D and VGG16-LSTM\@.
MMI presented an increment in InceptionV3-LSTM and a decrement in C3D-block-LSTM\@.
In general, for each dataset, there is no significant change when synthetically augmenting the RI\@.

For FT, SD impacts the performance of the models less.
CK+ only presented a significant increment in ResNet101-LSTM, ResNet18-LSTM, and ResNet3D-18.
For MMI, VGG16-LSTM decreased its accuracy, and no other method presented significant differences.
For MUG, ResNet3D-18 decreased its accuracy, and no other method presented significant differences.
OULU presented and incremented in InceptionV3-LSTM and a decrement in C3D\@.
In general, in MMI, the SD also decreased its accuracy significantly compared to DA and without augmentation.
And in MUG, no augmentation performed significantly better than the other two (\cf Table~\ref{app:tab:st-single-models-unified}).

When comparing DA \vs SD in RI, MMI did not present any significant difference while, for CK+, C3D, C3D-block-LSTM, and VGG16-LSTM presented a significant superiority of DA over SD\@.
For MUG, InceptionV3-LSTM and ResNet18-LSTM presented a superiority of SD over DA\@.
For OULU, C3D and VGG16-LSTM presented a superiority of DA over SD, while InceptionV3-LSTM, ResNet101-LSTM, and ResNet3D-101 presented the opposite.

In FT, only a few significant differences were noticed.
MUG did not present any significant improvement while, for CK+, ResNet18-LSTM showed significant superiority of SD over DA, and VGG16-LSTM presented the opposite.
For MMI, ResNet3D-18 presented a significant superiority of DA over SD\@.
For OULU, C3D presented a superiority of SD over DA, while C3D-block-LSTM presented the opposite.

Note that the classical data augmentation techniques (\ie, affine transformations) obtained the best result when selecting the maximum among the methods in most databases.
However, in general, with a $5\%$ significance level the FT methods outperformed them (\cf Table~\ref{app:tab:st-merged-single-models unified-dataset}).
Since the affine transformations explore different local features of the original data, it generalizes well to the changes in the current database, \ie, it creates an invariant model to the changes within the subjects in the database.
In contrast, the synthetic data generation explores new combinations within the database that generalize better, for instance, when trained and tested with diverse data (\cf Section~\ref{sec:merged-database}) or when tested on different data to evaluate the method's generalization capabilities (\cf Section~\ref{sec:cross-database}).
Due to the lack of challenging data within a single database, this benefit is not reflected in this experiment, but it becomes evident in the following experiments.
CK+ and OULU, respectively, the simplest and the most complex databases, presented significant impacts of using the data augmentation techniques.
Curiously, such impact was more evidenced in shallow models for CK+ and deep models for OULU.
We will resume this discussion in the following sections.

\begin{table}[tb]
  \centering
  \sisetup{
    table-format = 2.1,
    round-mode=places,
    round-precision=1,
    separate-uncertainty,
  }
  \caption{The literature methods' average accuracy and selected methods on our experiments on $5$-fold cross-validation on single- and merged-databases.  We highlight in bold the best results of each experiment set.}
  \label{tab:results:comparison}
  \scriptsize
  \setlength{\colsep}{6pt}
  \begin{tabular}{%
      l%
      S@{\hspace{\colsep}}%
      S@{\hspace{\colsep}}%
      S@{\hspace{\colsep}}%
      S@{\hspace{\colsep}}%
    }
    \toprule
    \textbf{Model} & \textbf{CK+} & \textbf{MMI} & \textbf{OULU} & \textbf{MUG} \\ %
    \midrule
    Liang \etal~\cite{Liang2019} & 99.6 & 80.7 & 91.0 & {--}\\ %
    Acharya \etal~\cite{Acharya2018} & {--} & {--} & {--} & {--} \\ %
    Kuo \etal~\cite{Kuo2018} & 98.4 & {--} & 91.6 & {--}\\ %
    Yan \etal~\cite{Yan2018} & 96.6 & {--} & {--} & {--} \\ %
    Ghimire \etal~\cite{Ghimire2017} & 97.80 & 77.22 & {--} & 95.50 \\ %
    Inception-ResNet-3D~\cite{Hasani2017} & 93.2  & 77.5 & {--} & {--} \\ %
    Kaya \etal~\cite{Kaya2017} & 98.32 & 70.31 & {--} & {--} \\ %
    MSCNN-PHRNN~\cite{Zhang2017} & 98.5 & 81.1 & 86.2 & {--}\\ %
    DTAGN~\cite{Jung2015} & 97.2 & 70.2 & 81.4 & {--}\\ %
    Aifanti \etal~\cite{Aifanti2014} & 94.31 & {--} & {--} & 92.76 \\ %
    3DCNN-DAP~\cite{Liu2014} & 92.4 & 63.4 & {--} & {--} \\ %
    \midrule
    \multicolumn{5}{c}{Single-Dataset $5$-Fold Cross-Validation}\\
    \midrule
    VGG16-LSTM-FT & 91.1375 & \bfseries 61.2 & 43.125 & 86.18311 \\ %
    VGG16-LSTM-FT-DA & \bfseries95.4 & 58.1521 & 75.0 & 88.1878 \\ %
    I3D-RI & 76.7459 & 31.4183 & \bfseries79.5 & 85.1081 \\ %
    I3D-FT-DS & 87.77156 & 44.49963925 & 71.04 & 85.0155672 \\ %
    ResNet3D-101-FT & 92.0466 & 53.8086 & 78.125 & \bfseries89.5 \\ %
    \midrule
    \multicolumn{5}{c}{Merged-Dataset $5$-Fold Cross-Validation}\\
    \midrule
    ResNet3D-101-FT & 94.158730 & 64.560786 & \bfseries81.0 & 89.706260  \\ %
    ResNet3D-101-FT-SD &  \bfseries95.1 & 65.920996 & 78.958333 & 88.372369  \\ %
    VGG16-LSTM-FT-SD & 93.841270 & \bfseries68.7 & 77.916667 &\bfseries90.9 \\ %
    \bottomrule
  \end{tabular}
\end{table}

\subsubsection{Comparison Against Existing Methods}
\label{sec:sota-comparison}

In Table~\ref{tab:results:comparison}, we present our highest results and existing experiments in the literature.
We show the detailed results on Table~\ref{app:tab:single-model} in Appendix~\ref{app:single-model}.
We highlight that our experiments are not directly comparable with those methods.
Even our single-database $k$-fold cross-validation, which is the closest to the reported methods, do not use the reported hyper-parameters for all the methods (see Table~\ref{tab:models:parameters}).
Recall that we synchronized the hyper-parameters and setups as much as possible to allow paired experiments.
Hence, we cannot consider our results a fair nor a direct comparison against the literature.%
\footnote{Note that it is common to see comparisons on the literature where the contrasted methods have a different set of parameters.  Nevertheless, these mistakes are rarely discussed.}
Nevertheless, we show the literature results as an upper bound to our metrics and contrast them.

As shown in Table~\ref{tab:results:comparison}, there is no consensus on the highest accuracy over the databases.
In general, it seems that initializing weights and fine-tuning them improves the results and that 2D based methods have the edge over the 3D ones for the CK+ and MMI databases.
While for OULU and MUG, the 3D based methods show the highest accuracy.
However, as discussed before in Section~\ref{sec:single-database}, these methods present high variability, drops in accuracy, and inconsistent results across databases.
Thus, a maximum value may be a misleading indicator.%
\footnote{A commonly reported indicator of the literature.}

In contrast to the literature, our best results are significantly below the maximum reported values~\cite{Liang2019, Ghimire2017, Kaya2017}.
\citepos{Liang2019} method uses a two-stream method to process the spatial and temporal information of the videos, with bidirectional recurrence.
Interestingly, they use classical data augmentation to enhance their results.
The work of \citet{Ghimire2017} relies on geometrical features with support vector machines and boosting to learn robust classifiers.
Moreover, \citepos{Kaya2017} work uses audiovisual information to produce deep and dense visual features with audio features fused with a kernel extreme learning machine.
However, these complex and ad-hoc networks were outside of the scope of our exploration.
Nevertheless, we find it interesting to show the wide range of results and lack of consistency in the existing approaches that lead to exhaustive searches to find local maxima that may not generalize well (as discussed in Section~\ref{sec:cross-database}).

\subsection{Merged-Dataset $k$-Fold Cross-Validation}
\label{sec:merged-database}

Another common approach to tackle limited data is to join existing databases together.
In this section, we analyze the effects of merging databases to increase the amount of data.

We created a new database by combining CK+, MMI, OULU-Casia, and MUG\@.
We created a $k$-fold cross-validation experiment where each $i$th fold is the union of the $i$th folds at each database when doing the $k$-fold cross-validation independently.
We took care of using the same folds of previous experiments to compare between merged and individual databases.
\Ie, each model-initialization--augmentation combination was trained once for each merged-fold of the $5$-fold cross-validation.
We obtained the test mean accuracy for each database so that each boxplot model-initialization--augmentation is comparable against its single-database counterpart.

In Fig.~\ref{fig:comparison-model-unified}, we observe the same general dominance of FT over RI in the single-database experiments.
MUG and OULU showed a significant dominance of RI with no augmentation and SD over FT with DA\@.
When using FT, all the datasets showed a significant dominance of no augmentation and SD over DA\@.
Similarly, when using RI, no augmentation dominates DA in all datasets and, for MMI, MUG and OULU, SD dominates DA\@.

In the right column of Fig.~\ref{fig:comparison}, we show the fine-grained average accuracy over the $5$-fold cross-validation per database (remember that the training was done over a merged-database).
We show the detailed experiments in Table~\ref{app:tab:merged-database}.

\subsubsection{Model Initialization}

Models such as C3D and VGG16-LSTM showed significant superiority of FT over RI, regardless of the data augmentation scheme.
C3D-block-LSTM and ResNet3D-101 evidenced this superiority for SD and no augmentation at all.
I3D and ResNet3D-18 experimented with no significant improvement of one initialization method \wrt the other, for any data augmentation scheme.
Besides these two methods, no significant difference when using DA was evidenced for I3D, InceptionV3-LSTM, and ResNet101-LSTM\@.
Only three cases reported superiority of RI over FT, all of them using DA: I3D on MUG, ResNet18-LSTM on MMI, and ResNet3D-18 on MUG\@.

In summary, the per-database improvement of FT over RI is significant (\cf Table~\ref{app:tab:st-merged-models-unified}).
And the former showed a decreased variance (\cf Fig.~\ref{fig:comparison-model-unified}).
And as a whole, the same trend was maintained (\cf Table~\ref{app:tab:st-merged-merged-models unified-dataset}).
However, when looked at a more fine-grained scale, we can see that the data augmentation technique dramatically affects the dominance of one initialization method over the other, and such behavior varies across the databases.

\subsubsection{Data Augmentation}

The experiments show a clear pattern related to the data augmentation method: DA decreases the models' accuracy, regardless of the initialization (\cf Fig.~\ref{fig:comparison-model-and-datasets-unified}).
Furthermore, SD increases the variance of the results.
The apparent increment in the accuracy of the models when using SD cannot be considered as significant (\cf Table~\ref{app:tab:st-merged-merged-models unified-dataset}).
When comparing DA against SD, the latter trivially attains a better performance.

We observed only three cases of improvement of accuracy, and all were only for SD\@.
VGG16-LSTM had an increment for RI and SD for MUG, and C3D-block-LSTM and ResNet3D-18 had an increment for FT for OULU\@.
The two models whose accuracy remained invariant the most when applying DA with RI are C3D and VGG16-LSTM\@.
However, this variance does not stand for FT, in which all the models significantly decreased their accuracy with DA\@.

It is interesting to notice that aggregating existing datasets---a somehow naive approach of augmenting data---may yield adverse outcomes due to the mixture and heterogeneity it induces.
Additional gaps in data space may be too much to be filled by the models while learning, as demonstrated by the experiments.
Augmentation techniques that traverse these data clusters may not be enough to close the gaps and instead hinder the overall accuracy.
In contrast, synthetic generation helps.
However, it is not a general solution since typical initialization yields lower results than the inductive bias in fine-tuned methods.
Moreover, notice that the aggregated results (see Table~\ref{app:tab:st-merged-merged-models unified-dataset}) cannot reject the hypothesis that each method and its synthetically augmented counterpart are equal.

\subsubsection{Comparison Against Single-database Training}
\label{sec:single-vs-merged}

This section compares the classification performance of the single-database training against the merged-database training on the same test sets.

Figure~\ref{fig:comparison-model-unified} shows that the use of DA and FT implies a significant drop in performance for the merged dataset \wrt the single one, regardless of the initialization.
RI also presents a drop in performance for the merged dataset, for CK+, MUG, and OULU, when using DA.
It can be seen that, in most cases, the effect of the merged database is detrimental to the performance.
However, if we compare RI on a single-database training with FT in the merged-database training and FT on a single-database training with RI in the merged-database training, it is evident that the configuration of FT attains the best results in both cases.
Another remarkable pattern is that, for FT, SD consistently attains better results than DA\@.
For RI, no pattern of this kind is observed.

In general (\cf Table~\ref{app:tab:st-merged-single-vs-merged unified-models and -dataset}), all the methods in the single database reject the hypothesis that any RI method in the merge database is greater than them.
In contrast, the FT methods on the merged database cannot reject the hypothesis that they are more significant than any single database method.
The exception to the trend is FD+DA in the merged database that fails to reject the hypothesis that it is greater than the augmented versions of RI in the single database.
Moreover, we notice an increase in the variance of the results in the different databases for at least one-third of the setups (\cf Fig.~\ref{fig:comparison-model-and-datasets-unified}).
In the merged database, particular exceptions for FT are C3D for MUG, ResNet3D-18 for MUG, and I3D for OULU\@.
In the first, the merged database increases performance; while it decreases performance in the other two cases.

We observed a different behavior for FT when using classical data augmentation.
For MUG and OULU, the merged database decreases the accuracy for all the models (\cf Fig.~\ref{fig:comparison-model-and-datasets-unified}), except for C3D, in which it significantly increases.
We observed no significant difference in CK+, except for InceptionV3-LSTM, in which the merged database decreased the performance.
Regarding MMI, the majority of models experimented with a decrease in accuracy, except for C3D, C3D-block-LSTM, and InceptionV3-LSTM, which presented no significant difference.
Notice that there is a large disparity between single and merged database results in the less constrained datasets.
In contrast, CK+ contains more homogeneous images that have a smaller disparity in the results.
Again, this reinforces the idea that heterogeneity in the data diminishes the gains when augmenting limited data problems.

Some models showed increments in their accuracy for FT with SD, such as C3D, C3D-block-LSTM, and ResNet3D-101.
MMI showed increments for C3D, I3D, ResNet18-LSTM, ResNet3D-101 and VGG16-LSTM\@.
In the rest of the cases, no significant difference was evidenced.
In the general case, the FT+SD for the merged database, we can reject the hypothesis that it is smaller or equal to other methods. 

When using random initialization, although the merged dataset's effect depends on the model and the dataset, most cases showed either a decrease or no significant difference.
Interestingly, random initializations do not take advantage of the augment in data (albeit with an increase in variance on the data), while the pre-trained models do.

As a remarkable pattern, we can say that MUG and OULU decreased their performance in general when using classical data augmentation, regardless of the initialization algorithm and the model used.

\subsection{Cross-Dataset Cross-Validation}
\label{sec:cross-database}

In this experiment, we did a cross-database evaluation, consisting of training on all the videos from one database and testing all the videos from all others.
This experiment's objective is to evaluate the generalization capabilities of the models when trained and tested on different domains.

Due to time and computational resources constraints, and based on the results obtained in the experiments presented previously, we decided to limit our cross-database experiments only to FT initialization since RI performed more poorly in all setups.
For this set of experiments, we incorporated a new variation in combining both DA and SD augmentation techniques.
We denote them as FT+DA+SD\@.

Additionally, for this generalization scenario, we decided to include the Acted Facial Expressions in the Wild~(AFEW) database~\cite{Dhall2011, Dhall2012}, which contains in-the-wild videos extracted from movies with seven spontaneous expressions.
Unlike the previous databases that have no partition for train, validation, or test, AFEW is divided into three groups of \num{773}, \num{383}, and \num{653} samples for, respectively, train, validation, and test.
This database is the most complex one, as it is the most heterogeneous.

\begin{figure*}[tb]
  \centering
  \resizebox{\linewidth}{!}{\begin{tikzpicture}
\begin{groupplot}[
  group style = {
    group size=5 by 1,
    horizontal sep = .4em,
    x descriptions at=edge bottom,
    y descriptions at=edge left,
  },
  width=0.5\linewidth,
  height=5cm,
  cycle list/Set3-4,
  boxplot/draw direction=y,
  enlarge x limits=.05,
  xmajorgrids,
  ylabel = {Accuracy (\%)},
  ymin = 0,
  ymax = 1,
  boxplot={
    draw position={1/5 + Floor(\plotnumofactualtype/4) + 1/5*mod(\plotnumofactualtype,4)},
    box extend=0.125,
  },
  xtick={0,1,2,...,5},
  x tick label as interval,
  x tick label style={
    align=center
  },
  legend entries = {{FT}, {FT+DA}, {FT+SD}, {FT+DA+SD}},
  legend to name={grouplegend},
  legend style={
    font=\footnotesize,
    legend columns=4,
    draw=none,
  },
]

\nextgroupplot[
xlabel={CK+},
  xticklabels = {MMI, OULU, MUG, AFEW},
]

\pgfplotsinvokeforeach{MMI, OULU, MUG, AFEW}{
  \def\dbname{#1}
  \pgfplotstableread[col sep=comma]{./tablesv2/cross_1vall/boxplot/CK/\dbname.csv}\csvdata
  \pgfplotstabletranspose\datatransposed{\csvdata}   
  
  \foreach \n in {1,...,4} {
    \addplot+[boxplot, fill, draw=black!50] table[y index=\n] {\datatransposed};
  }
}

\nextgroupplot[
  xlabel={MMI},
  xticklabels = {CK+, OULU, MUG, AFEW},
]

\pgfplotsinvokeforeach{CK, OULU, MUG, AFEW}{
  \def\dbname{#1}
  \pgfplotstableread[col sep=comma]{./tablesv2/cross_1vall/boxplot/MMI/\dbname.csv}\csvdata
  \pgfplotstabletranspose\datatransposed{\csvdata}   
  
  \foreach \n in {1,...,4} {
    \addplot+[boxplot, fill, draw=black!50] table[y index=\n] {\datatransposed};
  }
}

\nextgroupplot[
  xlabel={OULU},
  xticklabels = {CK+, MMI, MUG, AFEW},
]

\pgfplotsinvokeforeach{CK, MMI, MUG, AFEW}{
  \def\dbname{#1}
  \pgfplotstableread[col sep=comma]{./tablesv2/cross_1vall/boxplot/OULU/\dbname.csv}\csvdata
  \pgfplotstabletranspose\datatransposed{\csvdata}   
  
  \foreach \n in {1,...,4} {
    \addplot+[boxplot, fill, draw=black!50] table[y index=\n] {\datatransposed};
  }
}

\nextgroupplot[
  xlabel={MUG},
  xticklabels = {CK+, MMI, OULU, AFEW},
]

\pgfplotsinvokeforeach{CK, MMI, OULU, AFEW}{
	\def\dbname{#1}
	\pgfplotstableread[col sep=comma]{./tablesv2/cross_1vall/boxplot/MUG/\dbname.csv}\csvdata
	\pgfplotstabletranspose\datatransposed{\csvdata}   

	\foreach \n in {1,...,4} {
	  \addplot+[boxplot, fill, draw=black!50] table[y index=\n] {\datatransposed};
	}
}

\nextgroupplot[
xlabel={AFEW},
xticklabels = {CK+, MMI, OULU, MUG},
]

\pgfplotsinvokeforeach{CK, MMI, OULU, MUG}{
	\def\dbname{#1}
	\pgfplotstableread[col sep=comma]{./tablesv2/cross_1vall/boxplot/AFEW/\dbname.csv}\csvdata
	\pgfplotstabletranspose\datatransposed{\csvdata}   
	
	\foreach \n in {1,...,4} {
		\addplot+[boxplot, fill, draw=black!50] table[y index=\n] {\datatransposed};
	}
}

\end{groupplot}
\path (group c1r1.north east) -- node[above, yshift=0cm]{\pgfplotslegendfromname{grouplegend}} (group c5r1.north west);
\end{tikzpicture}}%
  \caption{
    The average accuracy of all the architectures on a cross-database evaluation.
    Each plot represents the training database, and the different groups are the accuracy when evaluated on that database.
    For initialization, we used transfer learning by fine-tuning previously trained models~(FT).
    We used affine transformations used in classical data augmentation~(DA) and synthetic data generated with models from the training partitions~(SD) for augmentation.
  }
  \label{fig:cross-database}
\end{figure*}
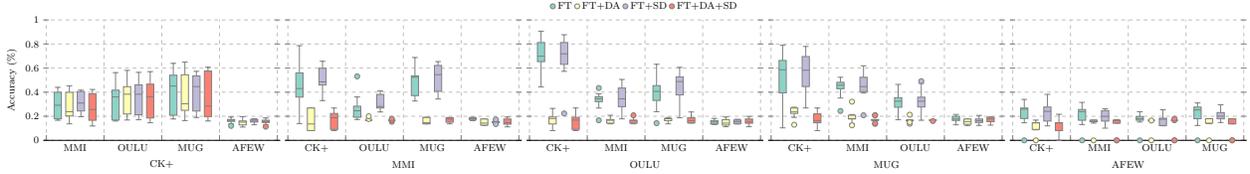

In our generalization test, we found that DA yielded lower accuracy \wrt SD\@.
See Fig.~\ref{fig:cross-database} for the database hold-out experiments.
We show the detailed results in Appendix~\ref{app:cross-databases}, where each table represents a model-initialization-augmentation combination.
Moreover, the addition of affine transformations to the synthetic data hurts the generalization capabilities of the models (\cf FT+DA+SD in Fig.~\ref{fig:cross-database}).

Interestingly, when training on a stable and straightforward database, like CK+, the methods' generalization is stable for the different augmentation techniques.
On the contrary, when training in more diverse databases, like MMI, OULU, and MUG, the affine data augmentation does not work.
In particular, when involving the AFEW database, either for training or testing, the attained results significantly dropped, regardless of the initialization and data augmentation strategy.
Note that even though using SD is minor when compared with FT alone, the variance is reduced \wrt the former.

\section{Discussion}
\label{sec:discussion}

In this section, we wrap up the comparisons of this study more conclusively by discussing the impact of each of the studied techniques.

\subsection{Impact of Initialization Techniques}

In this study, the most significant pattern observed is the superiority of inductive biases in the form of pre-trained models (\aka FT) over random initializations (RI).
The latter also presents higher variability in most of the cases for single-database training (Section~\ref{sec:single-database}).
We observed the dominance of FT over RI to happen regardless of both the datasets and the data augmentation techniques.
These two factors only affect the magnitude of such dominance.
Our experiments showed no impact on database stacking over the choice of initialization.
Furthermore, it showed clear superiority of FT with and without the dataset stacking.
Consequently, one may prefer to opt for fine-tuning instead of stacking databases when facing small data (Section~\ref{sec:merged-database}).
We also observed that the improvement provided by FT is more significant in deeper architectures when tested on a single database (Section~\ref{sec:single-database}).
RI also shows some detrimental effect on database stacking (Section~\ref{sec:merged-database}).

\subsection{Impact of Data Augmentation Techniques}

The type of data augmentation (either DA or SD) plays a vital role in the methods' performance.
We saw that it impacts the effect of database stacking, although to a lesser extent than initialization, since the effectiveness of the augmentation technique is tightly constrained by the initialization (Section~\ref{sec:merged-database}).
As an example, notice that, for RI, the effects of both augmentation techniques, DA and SD, are practically null, while, for FT, they are extremely heterogeneous (Section~\ref{sec:single-database}).
When training with a single database, the effects of DA are null for most of the datasets, but mainly on high-variance datasets like OULU and MMI\@.
These results suggest that merely affine transformations do not augment the diversity of such datasets.
When training on merged databases, DA generally decreases the performance of the models, while SD, besides not providing any significant improvement, increases the variance (Section~\ref{sec:merged-database}).
The only scenario in which SD significantly outperforms DA and no augmentation corresponds to the cross-database experiments.
This result shows that synthetic data augmentation is an effective technique to increase the models' generalization performance (Section~\ref{sec:cross-database}).

\subsection{Impact of Database Stacking}

Augmenting data by stacking datasets is not guaranteed to be an improvement (Section~\ref{sec:merged-database}).
The results we obtained in this study show that naively mixing databases without a care for their contents and similarities may produce sparse training sets that tax the model to the point to decrease its performance.
Simply said, if not carefully selected, the augmented data may increase the gaps of the base dataset instead of filling them.
We assume that the decrement in performance is related to the models being incapable of not bridging the training data gaps.
Notice that these gaps are not easily controllable without a care in understanding the data available.
Moreover, we suggest working with limited data and better inductive biases instead of blindly aggregating data.
And the final aggregated results (\cf Fig.~\ref{fig:comparison-model-and-datasets-unified}) support that fine-tuned methods outperform their counterparts.

\subsection{Impacts on Generalization through Cross-database Testing}

Our generalization studies by cross-database testing show that data augmentation through affine transformations works better within the same data setup but decay when testing in heterogeneous data setups (Section~\ref{sec:cross-database}).
That is, the affine transformation has limited generalization capabilities.
This behavior is explained by the fact that this technique only has affine transformations, \ie, it is only creating the same data presented in different shapes, which does not help the fit of the general sample space of real-world tasks.
This limitation results in a database over-fitting, responsible for the difference in metrics between training with only one database and training using several databases.

We found that using stable data (with fewer variations) produces more predictable and stable results over newer data regarding cross-database evaluation.
However, training in heterogeneous data, synthetic data, and fine-tuning helped improve performance in other scenarios, at the cost of a wider variability on the predicted performance.

\section{Conclusions}
\label{sec:conclusions}

We presented a large-scale study to show the improvements and limitations when training with limited data for classification problems dealing with facial expression recognition from video.
We explored the use of two model initialization techniques and two data augmentation methods, along with the possibility of stacking databases.
These variants were tested on nine widely-used deep architectures and in four datasets.
We performed an exhaustive analysis of all these variants and saw that the improvement in classification performance that these techniques provide is not straightforward, but full of nuances.
We showed how mixing these techniques yield different generalization levels and how significant the differences between them are in terms of classification performance.

Among the most overwhelming results obtained from the study, the dominance of inductive biases in the form of Fine-Tuning over other model initialization and data augmentation techniques stands out the most, suggesting that Fine-Tuning is the most likely technique to improve model performance when facing small datasets significantly.
A second insight from our study is that data augmentation, either classical or GAN-based, marginally improves most of the tested models' performance when no other technique (\eg database stacking or fine-tuning) is used.
When database stacking is used, GAN-based data augmentation is the most suited to fill the gaps created by merging heterogeneous data, while classical data augmentation falls short.
A third insight is that stacking heterogeneous databases is not a straightforward improvement, as the type of augmentation and initialization is crucial.
Besides that, the complexity of the test data and the complexity of the stacking datasets to train a model are essential aspects we must consider before merging data indiscriminately, as complex training sets translate better to more stable test sets when fine-tuned and coupled with GAN-based-synthetic data. On the other hand, more simple data translates to more stable performance on unseen data (albeit with lower performance).

\section*{Acknowledgments}
This project was funded by the Coordena\c{c}\~ao de Aperfei\c{c}oamento de Pessoal de N\'ivel Superior---Brasil (CAPES)---Finance Code 001; S\~ao Paulo Research Foundation (FAPESP) under grants No.~2016/19947-6, 2017/16144-2, and~2019/07257-3; and by the Brazilian National Council for Scientific and Technological Development (CNPq) under grant No.~307425/2017-7.

\printbibliography

\appendix

\makeatletter
\def\appendixname{}%
\gdef\thesection{\@Alph\c@section}%
\makeatother

\section{Models' Configuration}
\label{app:models}
\setcounter{table}{0}

In the tables of this section, we show the implementation details of each model used in the experiments.
The tables contain three columns: the name of the layer (or block), the output size, and the filter's configuration.

It should be noted that the configurations of models are the same as those introduced by the original authors; the only thing that changes is the elimination of the last layer (Softmax).
Then, in 2D convolutional-based networks, an LSTM recurrent block is added to the last layer, which has a depth-one layer.
Finally, in all models, a classifier block is stacked, composed of three dense layers that end in a softmax to make the prediction.

\bgroup
\renewcommand{\arraystretch}{1.5}
\begin{table}[tb]
	\centering
	\sisetup{
		table-format = 1,
	}
	\caption{Architecture details for VGG16~\cite{Simonyan2015} plus LSTM.}
	\label{app:models:vgg16}
	\scriptsize
	\setlength{\colsep}{4pt}
	\begin{tabular}{%
			l@{\hspace{\colsep}}%
			l@{\hspace{\colsep}}%
			l@{\hspace{\colsep}}%
		}
		\toprule
		\textbf{Layer Name} & \textbf{Output Size} & \textbf{Configuration} \\
		\midrule
		Conv1 & $224 \times 224 \times 64$  & $\begin{bmatrix}  3 \times 3,64 \\ 3 \times 3,64 \end{bmatrix} \times 2$\\
		Max pool & $112 \times 112 \times 64$ & {$2 \times 2$ max pool, stride 2}\\
		Conv2 & $112 \times 112 \times 128$  & {$\begin{bmatrix}  3 \times 3,128 \\ 3 \times 3,128 \end{bmatrix} \times 2$}\\
		Max pool & $56 \times 56 \times 128$ & {$2 \times 2$ max pool, stride 2}\\
		Conv3 & $56 \times 56 \times 256$  & {$\begin{bmatrix}  3 \times 3,256 \\ 3 \times 3,256 \end{bmatrix} \times 2$}\\
		Max pool & $28 \times 28 \times 256$ & {$2 \times 2$ max pool, stride 2}\\
		Conv4 & $28 \times 28 \times 512$  & {$\begin{bmatrix}  3 \times 3,512 \\ 3 \times 3,512 \end{bmatrix} \times 3$}\\
		Max pool & $14 \times 14 \times 512$ & {$2 \times 2$ max pool, stride 2}\\
		Conv5 & $14 \times 14 \times 512$  & {$\begin{bmatrix}  3 \times 3,512 \\ 3 \times 3,512 \end{bmatrix} \times 3$}\\
		Max pool & $7 \times 7 \times 512$ & {$2 \times 2$ max pool, stride 2}\\
		LSTM & $25088 \times 1024$ & {1 Layer} \\
		Fully connected & $512 \times 256$ & {$512 \times 256$  fully connections} \\
		Fully connected & $256 \times 128$ & {$256 \times 128$  fully connections} \\
		Fully connected & $128 \times N_{c}$ & {$512 \times N_{c}$  fully connections} \\
		Softmax & $N_{c}$ & {$N_{c}$: number of classes} \\
		
		\bottomrule
	\end{tabular}
\end{table}

\begin{table}[tb]
	\centering
	\sisetup{
		table-format = 1,
	}
	\caption{Architecture details for Inception V3~\cite{Szegedy2015, Szegedy2016, Szegedy2016a} plus LSTM.}
	\label{app:models:inceptionv3-lstm}
	\scriptsize
	\setlength{\colsep}{4pt}
	\begin{tabular}{%
			l@{\hspace{\colsep}}%
			l@{\hspace{\colsep}}%
			l@{\hspace{\colsep}}%
		}
		\toprule
		\textbf{Layer Name} & \textbf{Output Size} & \textbf{Configuration} \\
		\midrule
		Steam & $35 \times 35 \times 288$ & {$\begin{bmatrix}  
			3 \times 3,32\\
			3 \times 3,32\\
			3 \times 3,64\\
			3 \times 3,80\\
			3 \times 3,192\\
			3 \times 3,288\\
			\end{bmatrix} \times 1$}\\
		Inception A & $17 \times 17 \times 768$ & {$\begin{bmatrix}  
			1 \times 1 & 1 \times 1 & \text{Pool} & 1 \times 1\\
			3 \times 3 & 3 \times 3 & 1 \times 1 & {}\\
			3 \times 3 & \multicolumn{3}{c}{}\\
			\multicolumn{4}{c}{\text{Concat}} \\
			\end{bmatrix} \times 3$}\\
		Inception B & $8 \times 8 \times 1280$ & {$\begin{bmatrix}  
			1 \times 1 & 1 \times 1 & \text{Pool} & 1 \times 1\\
			1 \times 7 & 1 \times 7 & 1 \times 1 & {}\\
			7 \times 1 & 7 \times 1 & \multicolumn{2}{c}{}\\
			1 \times 7 & \multicolumn{3}{c}{}\\
			7 \times 1 & \multicolumn{3}{c}{}\\
			\multicolumn{4}{c}{\text{Concat}} \\
			\end{bmatrix} \times 3$}\\
		Inception C & $8 \times 8 \times 2048$ & {$\begin{bmatrix}  
			\multicolumn{2}{c}{1 \times 1} & \multicolumn{2}{c}{1 \times 1} & \text{Pool} & 1 \times 1\\
			\multicolumn{2}{c}{3 \times 3} & 1 \times 3 & 3 \times 1 & 1 \times 1 & \\
			1 \times 3 & 3 \times 1 & \multicolumn{4}{c}{}\\
			\multicolumn{6}{c}{\text{Concat}} \\
			\end{bmatrix} \times 2$}\\
		Max Pool & $1 \times 1 \times 2048$ & {$8 \times 8$ Max pool}\\
		LSTM & $2048 \times 512$ & {1 Layer} \\
		Fully connected & $512 \times 256$ & {$73728 \times 256$  fully connections} \\
		Fully connected & $256 \times 128$ & {$256 \times 128$  fully connections} \\
		Fully connected & $128 \times N_{c}$ & {$512 \times N_{c}$  fully connections} \\
		Softmax & $N_{c}$ & {$N_{c}$: number of classes} \\
		\bottomrule
	\end{tabular}
\end{table}

\begin{table}[tb]
	\centering
	\sisetup{
		table-format = 1,
	}
	\caption{Architecture details for ResNet-18~\cite{He2016} plus LSTM.}
	\label{app:models:resnet18}
	\scriptsize
	\setlength{\colsep}{4pt}
	\begin{tabular}{%
			l@{\hspace{\colsep}}%
			l@{\hspace{\colsep}}%
			l@{\hspace{\colsep}}%
		}
		\toprule
		\textbf{Layer Name} & \textbf{Output Size} & \textbf{Configuration} \\
		\midrule
		Conv1 & $112 \times 112 \times 64$ & {$7 \times 7$, 64, stride 2}\\
		Max pool & $56 \times 56 \times 64$ & $2 \times$ max pool, stride 2\\
		Conv2 & $56 \times 56$x 64  & {$\begin{bmatrix}  3 \times 3,64 \\ 3 \times 3,64 \end{bmatrix} \times 2$}\\
		Conv3 & $28 \times 28 \times 128$  & {$\begin{bmatrix}  3 \times 3,128 \\ 3 \times 3,128 \end{bmatrix} \times 2$}\\
		Conv4 & $14 \times 14 \times 256$  & {$\begin{bmatrix}  3 \times 3,256 \\ 3 \times 3,256 \end{bmatrix} \times 2$}\\
		Conv5 & $7 \times 7 \times 512$  & {$\begin{bmatrix}  3 \times 3,512 \\ 3 \times 3,512 \end{bmatrix} \times 2$}\\
		Average pool & $1 \times 1 \times 512$ & {$7 \times 7$ average pool} \\
		LSTM & $512 \times 512$ & {1 Layer} \\
		Fully connected & $512 \times 256$ & {$512 \times 256$  fully connections} \\
		Fully connected & $256 \times 128$ & {$256 \times 128$  fully connections} \\
		Fully connected & $128 \times N_{c}$ & {$512 \times N_{c}$  fully connections} \\
		Softmax & $N_{c}$ & {$N_{c}$: number of classes} \\
		
		\bottomrule
	\end{tabular}
\end{table}

\begin{table}[tb]
	\centering
	\sisetup{
		table-format = 1,
	}
	\caption{Architecture details for ResNet-101~\cite{He2016} plus LSTM.}
	\label{app:models:resnet101}
	\scriptsize
	\setlength{\colsep}{4pt}
	\begin{tabular}{%
			l@{\hspace{\colsep}}%
			l@{\hspace{\colsep}}%
			l@{\hspace{\colsep}}%
		}
		\toprule
		\textbf{Layer Name} & \textbf{Output Size} & \textbf{Configuration} \\
		\midrule
		Conv1 & $112 \times 112 \times 64$ & {$7 \times 7$, 64, stride 2}\\
		Max pool & $56 \times 56 \times 64$ & {$2 \times 2$ max pool, stride 2}\\
		Conv2 & $56 \times 56 \times 256$  & {$\begin{bmatrix}  1 \times 1,64 \\ 3 \times 3,64 \\ 1 \times 1,256 \end{bmatrix} \times 3$}\\
		Conv3 & $28 \times 28 \times 512$  & {$\begin{bmatrix}   1 \times 1,128 \\ 3 \times 3,128 \\ 1 \times 1,512 \end{bmatrix} \times 4$}\\
		Conv4 & $14 \times 14 \times 1024$  & {$\begin{bmatrix}   1 \times 1,256 \\ 3 \times 3,256 \\ 1 \times 1,1024 \end{bmatrix} \times 23$}\\
		Conv5 & $7 \times 7 \times 2048$  & {$\begin{bmatrix}   1 \times 1,512 \\ 3 \times 3,512 \\ 1 \times 1,2048 \end{bmatrix} \times 3$}\\
		Average pool & $1 \times 1 \times 2048$ & {$7 \times 7$ average pool} \\
		LSTM & $2048 \times 512$ & {1 Layer} \\
		Fully connected & $512 \times 256$ & {$512 \times 256$  fully connections} \\
		Fully connected & $256 \times 128$ & {$256 \times 128$  fully connections} \\
		Fully connected & $128 \times N_{c}$ & {$512 \times N_{c}$  fully connections} \\
		Softmax & $N_{c}$ & {$N_{c}$: number of classes} \\
		
		\bottomrule
	\end{tabular}
\end{table}

\begin{table}[tb]
	\centering
	\sisetup{
		table-format = 1,
	}
	\caption{Architecture details for C3D~\cite{Tran2015}.}
	\label{app:models:c3d}
	\scriptsize
	\setlength{\colsep}{4pt}
	\begin{tabular}{%
			l@{\hspace{\colsep}}%
			l@{\hspace{\colsep}}%
			l@{\hspace{\colsep}}%
		}
		\toprule
		\textbf{Layer Name} & \textbf{Output Size} & \textbf{Configuration} \\
		\midrule
		Conv1 & $100 \times 100 \times 16 \times 64$  & {$\begin{bmatrix}  3 \times 3 \times 3,64  \end{bmatrix} \times 1$}\\
		Max pool & $50 \times 50 \times 16 \times 64$& {$2 \times 2 \times 1$ max pool, stride 2}\\
		Conv2 & $50 \times 50 \times 16 \times 128$  & {$\begin{bmatrix}  3 \times 3 \times 3,128  \end{bmatrix} \times 1$}\\
		Max pool & $25 \times 25 \times 16 \times 128$ & {$2 \times 2 \times 1$ max pool, stride 2}\\
		Conv3 & $25 \times 25 \times 16 \times 256$  & {$\begin{bmatrix}  3 \times 3 \times 3,256  \end{bmatrix} \times 2$}\\
		Max pool & $12 \times 12 \times 16 \times 256$ & {$2 \times 2 \times 1$ max pool, stride 2}\\
		Conv4 & $12 \times 12 \times 16 \times 512$  & {$\begin{bmatrix}  3 \times 3 \times 3,512  \end{bmatrix} \times 2$}\\
		Max pool & $6 \times 6 \times 16 \times 512$ & {$2 \times 2 \times 1$ max pool, stride 2}\\
		Conv5 & $6 \times 6 \times 16 \times 512$  & {$\begin{bmatrix}  3 \times 3 \times 3,512  \end{bmatrix} \times 2$}\\
		Max pool & $3 \times 3 \times 16 \times 512$ & {$2 \times 2 \times 1$ max pool, stride 2}\\
		Fully connected & $73728 \times 256$ & {$73728 \times 256$  fully connections} \\
		Fully connected & $256 \times 128$ & {$256 \times 128$  fully connections} \\
		Fully connected & $128 \times N_{c}$ & {$512 \times N_{c}$  fully connections} \\
		Softmax & $N_{c}$ & {$N_{c}$: number of classes} \\
		
		\bottomrule
	\end{tabular}
\end{table}

\begin{table}[tb]
	\centering
	\sisetup{
		table-format = 1,
	}
	\caption{Architecture details for I3D~\cite{Carreira2017}.}
	\label{app:models:I3D}
	\scriptsize
	\setlength{\colsep}{4pt}
	\begin{tabular}{%
			l@{\hspace{\colsep}}%
			l@{\hspace{\colsep}}%
			l@{\hspace{\colsep}}%
		}
		\toprule
		\textbf{Layer Name} & \textbf{Output Size} & \textbf{Configuration} \\
		\midrule
		Conv 1 & $112 \times 112 \times 32 \times 64$ & {$\begin{bmatrix}  
			7 \times 7 \times 7, 64 \\
			\end{bmatrix} \times 1$}\\
		Max Pool & $56 \times 56 \times 16 \times 64$ &  {$2 \times 2 \times 2$~Max~pool,~stride~2}\\
		Conv 2 & $56 \times 56 \times 16 \times 192$ & {$\begin{bmatrix}  
			3 \times 3 \times 3, 192 \\
			\end{bmatrix} \times 1$}\\
		Max Pool & $28 \times 28 \times 8 \times 192$ &  {$2 \times 2 \times 2$~Max~pool,~stride~2}\\
		Inception 3a & $28 \times 28 \times 8 \times 256$ & {$\begin{bmatrix}  
			1 \times 1 \times 1 & 1 \times 1 \times 1 & 1 \times 1 \times 1 & \text{Max Pool}\\
			& 3 \times 3 \times 3 & 3 \times 3 \times 3 & 1 \times 1 \times 1\\
			\multicolumn{4}{c}{\text{Concat}} \\
			\end{bmatrix} \times 1$}\\
		Inception 3b & $28 \times 28 \times 8 \times 480$ & {$\begin{bmatrix}  
			1 \times 1 \times 1 & 1 \times 1 \times 1 & 1 \times 1 \times 1 & \text{Max Pool}\\
			& 3 \times 3 \times 3 & 3 \times 3 \times 3 & 1 \times 1 \times 1\\
			\multicolumn{4}{c}{\text{Concat}} \\
			\end{bmatrix} \times 1$}\\
		Max Pool & $14 \times 14 \times 4 \times 480$ &  {$2 \times 2 \times 2$~Max~pool,~stride~2}\\
		Inception 4a & $14 \times 14 \times 4 \times 512$ & {$\begin{bmatrix}  
			1 \times 1 \times 1 & 1 \times 1 \times 1 & 1 \times 1 \times 1 & \text{Max Pool}\\
			& 3 \times 3 \times 3 & 3 \times 3 \times 3 & 1 \times 1 \times 1\\
			\multicolumn{4}{c}{\text{Concat}} \\
			\end{bmatrix} \times 1$}\\
		Inception 4b & $14 \times 14 \times 4 \times 512$ & {$\begin{bmatrix}  
			1 \times 1 \times 1 & 1 \times 1 \times 1 & 1 \times 1 \times 1 & \text{Max Pool}\\
			& 3 \times 3 \times 3 & 3 \times 3 \times 3 & 1 \times 1 \times 1\\
			\multicolumn{4}{c}{\text{Concat}} \\
			\end{bmatrix} \times 1$}\\
		Inception 4c & $14 \times 14 \times 4 \times 512$ & {$\begin{bmatrix}  
			1 \times 1 \times 1 & 1 \times 1 \times 1 & 1 \times 1 \times 1 & \text{Max Pool}\\
			& 3 \times 3 \times 3 & 3 \times 3 \times 3 & 1 \times 1 \times 1\\
			\multicolumn{4}{c}{\text{Concat}} \\
			\end{bmatrix} \times 1$}\\
		Inception 4d & $14 \times 14 \times 4 \times 528$ & {$\begin{bmatrix}  
			1 \times 1 \times 1 & 1 \times 1 \times 1 & 1 \times 1 \times 1 & \text{Max Pool}\\
			& 3 \times 3 \times 3 & 3 \times 3 \times 3 & 1 \times 1 \times 1\\
			\multicolumn{4}{c}{\text{Concat}} \\
			\end{bmatrix} \times 1$}\\
		Inception 4e & $14 \times 14 \times 4 \times 832$ & {$\begin{bmatrix}  
			1 \times 1 \times 1 & 1 \times 1 \times 1 & 1 \times 1 \times 1 & \text{Max Pool}\\
			& 3 \times 3 \times 3 & 3 \times 3 \times 3 & 1 \times 1 \times 1\\
			\multicolumn{4}{c}{\text{Concat}} \\
			\end{bmatrix} \times 1$}\\
		Max Pool & $7 \times 7 \times 2 \times 832$ &  {$2 \times 2 \times 2$~Max~pool,~stride~2}\\
		Inception 5a & $7 \times 7 \times 2 \times 832$ & {$\begin{bmatrix}  
			1 \times 1 \times 1 & 1 \times 1 \times 1 & 1 \times 1 \times 1 & \text{Max Pool}\\
			& 3 \times 3 \times 3 & 3 \times 3 \times 3 & 1 \times 1 \times 1\\
			\multicolumn{4}{c}{\text{Concat}} \\
			\end{bmatrix} \times 1$}\\
		Inception 5b & $7 \times 7 \times 2 \times 1024$ & {$\begin{bmatrix}  
			1 \times 1 \times 1 & 1 \times 1 \times 1 & 1 \times 1 \times 1 & \text{Max Pool}\\
			& 3 \times 3 \times 3 & 3 \times 3 \times 3 & 1 \times 1 \times 1\\
			\multicolumn{4}{c}{\text{Concat}} \\
			\end{bmatrix} \times 1$}\\
		Average Pool & $1 \times 1 \times 1 \times 1024$ & {$7 \times 7 \times 2$ Max pool}\\
		Fully connected & $1024 \times 256$ & {$73728 \times 256$  fully connections} \\
		Fully connected & $256 \times 128$ & {$256 \times 128$  fully connections} \\
		Fully connected & $128 \times N_{c}$ & {$512 \times N_{c}$  fully connections} \\
		Softmax & $N_{c}$ & {$N_{c}$: number of classes} \\
		\bottomrule
	\end{tabular}
\end{table}

\begin{table}[tb]
	\centering
	\sisetup{
		table-format = 1,
	}
	\caption{Architecture details for ResNet3D-18~\cite{Hara2018}.}
	\label{app:models:resnet3d18}
	\scriptsize
	\setlength{\colsep}{4pt}
	\begin{tabular}{%
			l@{\hspace{\colsep}}%
			l@{\hspace{\colsep}}%
			l@{\hspace{\colsep}}%
		}
		\toprule
		\textbf{Layer Name} & \textbf{Output Size} & \textbf{Configuration} \\
		\midrule
		Conv1 & $100 \times 100 \times 25 \times 64$ & {$5 \times 5 \times 5$, 64, stride 2}\\
		Max3D pool & $50 \times 50 \times 12 \times 64$ & {$2 \times 2 \times 2$ max pool, stride 2}\\
		Conv2 & $50 \times 50 \times 12 \times 64$  & {$\begin{bmatrix}  3 \times 3 \times 1,64 \\ 3 \times 3 \times 1,64 \end{bmatrix} \times 2$}\\
		Conv3 & $25 \times 25 \times 6 \times 128$  & {$\begin{bmatrix}  3 \times 3 \times 1,128 \\ 3 \times 3 \times 1,128 \end{bmatrix} \times 2$}\\
		Conv4 & $12 \times 12 \times 3 \times 256$  & {$\begin{bmatrix}  3 \times 3 \times 3,256 \\ 3 \times 3 \times 3,256 \end{bmatrix} \times 2$}\\
		Conv5 & $6 \times 6 \times 1 \times 512$  & {$\begin{bmatrix}  3 \times 3 \times 3,512 \\ 3 \times 3 \times 3,512 \end{bmatrix} \times 2$}\\
		Average3D pool & $1 \times 1 \times 1 \times 512$ & {$6 \times 6$ average pool} \\
		Fully connected & $512 \times 256$ & {$512 \times 256$  fully connections} \\
		Fully connected & $256 \times 128$ & {$256 \times 128$  fully connections} \\
		Fully connected & $128 \times N_{c}$ & {$512 \times N_{c}$  fully connections} \\
		Softmax & $N_{c}$ & {$N_{c}$: number of classes} \\
		
		\bottomrule
	\end{tabular}
\end{table}

\begin{table}[tb]
	\centering
	\sisetup{
		table-format = 1,
	}
	\caption{Architecture details for ResNet3D-101~\cite{Hara2018}.}
	\label{app:models:resnet3d101}
	\scriptsize
	\setlength{\colsep}{4pt}
	\begin{tabular}{%
			l@{\hspace{\colsep}}%
			l@{\hspace{\colsep}}%
			l@{\hspace{\colsep}}%
		}
		\toprule
		\textbf{Layer Name} & \textbf{Output Size} & \textbf{Configuration} \\
		\midrule
		Conv1 & $100 \times 100 \times 25 \times 64$ & {$\begin{bmatrix}  7 \times 7 \times 7,64\\ \end{bmatrix} \times 1$}\\
		Max3D pool & $50 \times 50 \times 12 \times 64$ & {$2 \times 2 \times 2$ max pool, stride 2}\\
		Conv2 & $50 \times 50 \times 6 \times 256$  & {$\begin{bmatrix}  1 \times 1 \times 1,64 \\ 3 \times  3 \times 3,64 \\ 1 \times 1 \times 1,256 \end{bmatrix} \times 3$}\\
		Conv3 & $25 \times 25 \times 3 \times 512$  & {$\begin{bmatrix}   1 \times 1 \times 1,128 \\ 3 \times 3 \times 3,128 \\ 1 \times 1 \times 1,512 \end{bmatrix} \times 4$}\\
		Conv4 & $12 \times 12 \times 1 \times 1024$  & {$\begin{bmatrix}   1 \times 1 \times 1,256 \\ 3 \times 3 \times 3,256 \\ 1 \times 1 \times 1,1024 \end{bmatrix} \times 23$}\\
		Conv5 & $6 \times 6 \times 1 \times 2048$  & {$\begin{bmatrix}   1 \times 1 \times 1,512 \\ 3 \times 3 \times 3,512 \\ 1 \times 1 \times 1,2048 \end{bmatrix} \times 3$}\\
		Average3D pool & $1 \times 1 \times 1 \times 2048$ & {$6 \times 6 \times 1$ average pool} \\
		Fully connected & $2048 \times 256$ & {$512 \times 256$  fully connections} \\
		Fully connected & $256 \times 128$ & {$256 \times 128$  fully connections} \\
		Fully connected & $128 \times N_{c}$ & {$512 \times N_{c}$  fully connections} \\
		Softmax & $N_{c}$ & {$N_{c}$: number of classes} \\
		
		\bottomrule
	\end{tabular}
\end{table}

\begin{table}[tb]
	\centering
	\sisetup{
		table-format = 1,
	}
	\caption{Architecture details for C3D-block-LSTM.}
	\label{app:models:c3d-block-lstm}
	\scriptsize
	\setlength{\colsep}{4pt}
	\begin{tabular}{%
			l@{\hspace{\colsep}}%
			l@{\hspace{\colsep}}%
			l@{\hspace{\colsep}}%
		}
		\toprule
		\textbf{Layer Name} & \textbf{Output Size} & \textbf{Configuration} \\
		\midrule
		Conv1 & $100 \times 100 \times 16 \times 64$  & {$\begin{bmatrix}  3 \times 3 \times 3,64  \end{bmatrix} \times 1$}\\
		Max pool & $50 \times 50 \times 16 \times 64$& {$2 \times 2 \times 1$ max pool, stride 2}\\
		Conv2 & $50 \times 50 \times 16 \times 128$  & {$\begin{bmatrix}  3 \times 3 \times 3,128  \end{bmatrix} \times 1$}\\
		Max pool & $25 \times 25 \times 16 \times 128$ & {$2 \times 2 \times 1$ max pool, stride 2}\\
		Conv3 & $25 \times 25 \times 16 \times 256$  & {$\begin{bmatrix}  3 \times 3 \times 3,256  \end{bmatrix} \times 2$}\\
		Max pool & $12 \times 12 \times 16 \times 256$ & {$2 \times 2 \times 1$ max pool, stride 2}\\
		Conv4 & $12 \times 12 \times 16 \times 512$  & {$\begin{bmatrix}  3 \times 3 \times 3,512  \end{bmatrix} \times 2$}\\
		Max pool & $6 \times 6 \times 16 \times 512$ & {$2 \times 2 \times 1$ max pool, stride 2}\\
		Conv5 & $6 \times 6 \times 16 \times 512$  & {$\begin{bmatrix}  3 \times 3 \times 3,512  \end{bmatrix} \times 2$}\\
		Max pool & $3 \times 3 \times 16 \times 512$ & {$2 \times 2 \times 1$ max pool, stride 2}\\
		LSTM & $73728 \times 512$ & {1 Layer} \\
		Fully connected & $512 \times 256$ & {$73728 \times 256$  fully connections} \\
		Fully connected & $256 \times 128$ & {$256 \times 128$  fully connections} \\
		Fully connected & $128 \times N_{c}$ & {$512 \times N_{c}$  fully connections} \\
		Softmax & $N_{c}$ & {$N_{c}$: number of classes} \\
		
		\bottomrule
	\end{tabular}
\end{table}
\egroup
\section{Single-Dataset  $k$-Fold Cross-Validation Ablation Results}
\label{app:single-model}
\setcounter{table}{0}

We show the detailed results of the experiments of all the methods when trained and tested on the same database in Table~\ref{app:tab:single-model}.
The configurations of each method correspond to it with and without fine-tuning and with different data augmentation.

We include results on the AFEW dataset (introduced in Section~\ref{sec:cross-database}) for the sigle-, merged-, and cross-dataset scenarios.
However, it is essential to remark that we kept the original partition of this dataset and did not perform 5-fold cross-validation experiments. 
The results reported in this and the following sections involving AFEW are not directly comparable to the other datasets for the single- and merged-dataset experiments.

\begin{filecontents*}{./tablesv2/metadata/models.csv}
modelname,rendername
VGG16-lstm,VGG16-LSTM
Inceptionv3-lstm,InceptionV3-LSTM
ResNet18-lstm,ResNet18-LSTM
ResNet101-lstm,ResNet101-LSTM
C3D,C3D
c3d-block-lstm,C3D-Block-LSTM
I3D,I3D
ResNet3D-18,ResNet3D-18
ResNet3D-101,ResNet3D-101
\end{filecontents*}
\DTLloaddb[noheader=false]{ablation-kfold-metadata}{./tablesv2/metadata/models.csv}
\begin{table}[tb]
\centering
\sisetup{
  table-format = 1.3(2),
  round-mode=places,
  round-precision=3,
  separate-uncertainty,
}
\caption{Ablation study of several methods when trained and tested on the \textbf{same database} with random initialization~(RI) or fine-tuning~(FT), \ie, pre-training and then transfer learning, and with data augmentation using random transformations~(DA), using synthetic generated data~(SD), and without it~(--).}
\label{app:tab:single-model}
\scriptsize
\resizebox{\linewidth}{!}{%
\begin{tabular}{
  l%
  S%
  S%
  S%
  S%
  S%
  S%
  S[table-format=1.3]%
}
  \toprule
  \multirow{2}{*}{\textbf{Architecture}} & \multicolumn{2}{c}{\textbf{Techniques}} & \multicolumn{5}{c}{\textbf{Datasets}} \\
  \cmidrule{2-8}
  & \textbf{Init.} & \textbf{Data Aug.} & \textbf{CK+} & \textbf{MMI}  & \textbf{OULU} & \textbf{MUG} & \textbf{AFEW} \\
  \midrule
\DTLforeach{ablation-kfold-metadata}{\modelname=modelname,\rendername=rendername}{%
  \csvreader[head to column names]{./tablesv2/auto_ablation/models/\modelname.csv}{}{%
   \ifnumequal{\thecsvrow}{1}{{\rendername}}{} & %
   {\expandafter\ifdefstring\expandafter{\ft}{1}{FT}{RI}} & %
   {\expandafter\ifdefstring\expandafter{\da}{R}{DA}{\expandafter\ifdefstring\expandafter{\da}{S}{SD}{--}}} &
    \ckmean(\ckstd) & \mmimean(\mmistd) & \oulumean(\oulustd) & \mugmean(\mugstd)  & \afewmean \\
   \ifnumequal{\thecsvrow}{6}{\midrule}{}
  }%
}%
\end{tabular}
}
\end{table}
\section{Merged-Dataset  $k$-Fold Cross-Validation Ablation Results}
\label{app:merged-database}
\setcounter{table}{0}

We show detailed results of the experiments of all the methods when trained on the merged folds of all databases and evaluated in each database individually in Table~\ref{app:tab:merged-database}.
The configurations of each method correspond to it with and without fine-tuning and with different data augmentation.

\DTLloaddb[noheader=false]{cross-ablation-kfold-metadata}{./tablesv2/metadata/models.csv}

\begin{table}[tb]
\centering
\sisetup{
  table-format = 1.3(2),
  round-mode=places,
  round-precision=3,
  separate-uncertainty,
}
\caption{Ablation study of several methods when trained and tested on the \textbf{combination of databases} with random initialization~(RI) or fine-tuning~(FT), \ie, pre-training and then transfer learning, and with data augmentation using random transformations~(DA), using synthetic generated data~(SD), and without it~(--).}
\label{app:tab:merged-database}
\scriptsize
\resizebox{\linewidth}{!}{%
\begin{tabular}{
  l%
  S%
  S%
  S%
  S%
  S%
  S%
  S%
  }
  \toprule
  \multirow{2}{*}{\textbf{Architecture}} & \multicolumn{2}{c}{\textbf{Techniques}} & \multicolumn{5}{c}{\textbf{Datasets}}\\
  \cmidrule{2-8}
  & \textbf{Init.} & \textbf{Data Aug.} & \textbf{CK+} & \textbf{MMI}  & \textbf{OULU} & \textbf{MUG} & \textbf{All} \\
  \midrule
  \DTLforeach*{cross-ablation-kfold-metadata}{\modelname=modelname,\rendername=rendername}{%
    \csvreader[head to column names]{./tablesv2/cross_kfold_ablation/models/\modelname.csv}{}{%
      \ifnumequal{\thecsvrow}{1}{{\rendername}}{} & %
      {\expandafter\ifdefstring\expandafter{\ft}{1}{FT}{RI}} & %
      {\expandafter\ifdefstring\expandafter{\da}{R}{DA}{\expandafter\ifdefstring\expandafter{\da}{S}{SD}{--}}} &
      \ckmean(\ckstd) & \mmimean(\mmistd) & \oulumean(\oulustd) & \mugmean(\mugstd)  & \meanmean(\meanstd) \\
      \ifnumequal{\thecsvrow}{6}{\midrule}{}
    }
  }
\end{tabular}
}
\end{table}

\section{Cross-Dataset Versus All Results}
\label{app:cross-databases}
\setcounter{table}{0}
\newcommand{\cmp}{--}

Each table shows the results of a model-initialization-augmentation combination.
Due to time and computational resource constraints, we decided to experiment with fine-tuning as our initialization scheme since the random one yielded lower performance in comparison.

\DTLloaddb[noheader=false]{appendix-cross1vall-models-metadata}{./tablesv2/metadata/models.csv}

\begin{filecontents*}{./tablesv2/metadata/config.csv}
config,cname
1000,Fine Tuning
1001,Fine Tuning + Data augmentation
1010,Fine Tuning + Synthetic augmentation
1011,Fine Tuning + Data augmentation + Synthetic augmentation
\end{filecontents*}
\DTLloaddb[noheader=false]{appendix-cross1vall-configs-metadata}{./tablesv2/metadata/config.csv}

\DTLforeach*{appendix-cross1vall-models-metadata}{\modelname=modelname, \rendername=rendername}{
  \DTLforeach*{appendix-cross1vall-configs-metadata}{\config=config,\cname=cname}{
  \begin{table}[tb]
    \centering
    \sisetup{
      table-format = 1.3,
      round-mode=places,
      round-precision=3,
      separate-uncertainty,
    }
    \scriptsize
    \caption{Classification accuracies of \rendername\ with \cname\ when trained in one database and evaluated in others.}
    \label{app:tab:cd-\modelname-\config}
    \setlength{\colsep}{6pt}
    \begin{tabular}{%
        l%
        S%
        S%
        S%
        S%
        S%
      }
      \toprule
      \textbf{Train} & \textbf{CK+} & \textbf{MMI} & \textbf{OULU}  & \textbf{MUG} & \textbf{AFEW}\\
      \midrule
      \csvreader[head to column names]{./tablesv2/cross_1vall/tables/\modelname/\config.csv}{}{
        \edef\tmp{\db}\ifdefstring{\tmp}{CK}{CK+}{\tmp} &
        \edef\tmp{\ck}\edef\nck{\num{\ck}}\ifdefstring{\tmp}{-}{\cmp}{\nck} &
        \edef\tmp{\mmi}\edef\nmmi{\num{\mmi}}\ifdefstring{\tmp}{-}{\cmp}{\nmmi} &
        \edef\tmp{\oulu}\edef\noulu{\num{\oulu}}\ifdefstring{\tmp}{-}{\cmp}{\noulu} &
        \edef\tmp{\mug}\edef\nmug{\num{\mug}}\ifdefstring{\tmp}{-}{\cmp}{\nmug} &
        \edef\tmp{\afew}\edef\nafew{\num{\afew}}\ifdefstring{\tmp}{-}{\cmp}{\nafew} \\
        \ifnumequal{\thecsvrow}{5}{\bottomrule}{}
      }
    \end{tabular}
  \end{table}
  }
}
\section{Statistical Hypothesis Tests for Experiments}
\label{app:stat-test}
\setcounter{table}{0}

In order to compare the results of our different experiments, we performed a Wilcoxon Signed-Rank Test.  
As suggested in the literature~\cite{Dietterich1998, Demvsar2006}, more thorough comparisons are needed to conclude the similarity between methods.  
However, this comparison is commonly ignored due to the lack of information on used folds in different databases.

Since we are constructing our experiments, we took care of using the same folds for all experiments.
Hence, we can consider two methods running on the same folds a paired experiment, and, thus, we used the Wilcoxon Signed-Rank test (a non-parametric test for comparison)~\cite{Demvsar2006}.

In the the tables of this section we present the $p$-value of each test pair-wise per database for the single- and merged-database (as explained in Sections~\ref{sec:single-database} and~\ref{sec:merged-database}, respectively).
We show the statistical test at two- and one-tail.
The former evaluates whether the methods' accuracy is equal, while the former evaluate if one is greater.
We denote by shading the $p$-values the rejection of the null hypothesis for each test.

\pgfplotstableset{
multistyler/.style 2 args={
  @my multistyler/.style={display columns/##1/.append style={#2}},
  @my multistyler/.list={#1}
},
hypothesis table/.style={
  col sep=comma,
  empty cells with={--}, %
  every column/.style={
    fixed, zerofill,
    precision=2,
  },
  multistyler={2,...,56}{%
    postproc cell content/.append code={%
      \ifstrequal{########1}{}{}{%
        \pgfmathparse{int(less(########1, 0.05))}%
        \ifnum\pgfmathresult=1%
        \pgfkeysalso{/pgfplots/table/@cell content/.add={\cellcolor{lightgray}}{}}%
        \fi%
      }%
    },
  },
  columns/models/.style={
    string type,
    column type=l,
    string replace={VGG16-lstm}{VGG16-LSTM},
    string replace={Inceptionv3-lstm}{InceptionV3-LSTM},
    string replace={ResNet18-lstm}{ResNet18-LSTM},
    string replace={ResNet101-lstm}{ResNet101-LSTM},
    string replace={c3d-block-lstm}{C3D-Block-LSTM},
    assign cell content/.code={
      \pgfmathparse{int(Mod(\pgfplotstablerow,6))}%
      \ifnum\pgfmathresult=0%
      \pgfkeyssetvalue{/pgfplots/table/@cell content}{####1}%
      \else%
      \pgfkeyssetvalue{/pgfplots/table/@cell content}{ }%
      \fi%
    },
  },
  columns/init/.style={
    string type,
    column type=l,
    string replace={0000}{RI},
    string replace={0001}{RI+SD},
    string replace={0010}{RI+DA},
    string replace={1000}{FT},
    string replace={1001}{FT+SD},
    string replace={1010}{FT+DA},
  },
  every head row/.style={
    output empty row,
    before row={
      \toprule%
      && \multicolumn{6}{l}{VGG16-LSTM} & \multicolumn{6}{l}{InceptionV3-LSTM} & \multicolumn{6}{l}{ResNet18-LSTM} & \multicolumn{6}{l}{ResNet101-LSTM} & \multicolumn{6}{l}{C3D} & \multicolumn{6}{l}{C3D-Block-LSTM} & \multicolumn{6}{l}{I3D} & \multicolumn{6}{l}{ResNet3D-18} & \multicolumn{6}{l}{ResNet3D-101} \\%
      \cmidrule{3-56}%
      && \footnotesize RI & \footnotesize RI+DA & \footnotesize RI+SD & \footnotesize FT & \footnotesize FT+DA & \footnotesize FT+SD 
      & \footnotesize RI & \footnotesize RI+DA & \footnotesize RI+SD & \footnotesize FT & \footnotesize FT+DA & \footnotesize FT+SD 
      & \footnotesize RI & \footnotesize RI+DA & \footnotesize RI+SD & \footnotesize FT & \footnotesize FT+DA & \footnotesize FT+SD 
      & \footnotesize RI & \footnotesize RI+DA & \footnotesize RI+SD & \footnotesize FT & \footnotesize FT+DA & \footnotesize FT+SD 
      & \footnotesize RI & \footnotesize RI+DA & \footnotesize RI+SD & \footnotesize FT & \footnotesize FT+DA & \footnotesize FT+SD 
      & \footnotesize RI & \footnotesize RI+DA & \footnotesize RI+SD & \footnotesize FT & \footnotesize FT+DA & \footnotesize FT+SD 
      & \footnotesize RI & \footnotesize RI+DA & \footnotesize RI+SD & \footnotesize FT & \footnotesize FT+DA & \footnotesize FT+SD 
      & \footnotesize RI & \footnotesize RI+DA & \footnotesize RI+SD & \footnotesize FT & \footnotesize FT+DA & \footnotesize FT+SD 
      & \footnotesize RI & \footnotesize RI+DA & \footnotesize RI+SD & \footnotesize FT & \footnotesize FT+DA & \footnotesize FT+SD \\%
    },
    after row=\midrule,
  },
  every last row/.style={after row=\bottomrule},
  every even row/.style={
    before row={\rowcolor[gray]{0.95}}
  },
},
}
\newcommand{\typesettable}[1]{%
\resizebox{%
\linewidth
}{!}{%
\pgfplotstabletypeset[%
  hypothesis table,
]{#1}}
}

\foreach \dbname in {CK, MMI, OULU, MUG}{%
  \ifdefstring{\dbname}{CK}{\def\dbnameshow{CK+}}{\def\dbnameshow{\dbname}}%
  \foreach \folder/\type in {auto_ablation/single,cross_kfold_ablation/merged,mix/single-vs-merged}{%
    \begin{landscape}%
      \begin{table}%
        \centering%
        \caption{$p$-values of the hypothesis testing for (a)~$H_0: M_1 = M_2$, $H_a: M_1 \neq M_2$ and (b)~$H_0: M_1 \leq M_2$, $H_a: M_1 > M_2$ on \dbnameshow\ for \type-database accuracies, where $M_1$ are the methods on the left and $M_2$ are the methods on top.  The shaded cells denote the rejection of the null hypothesis at $5\%$ significance level.}%
        \label{app:tab:st-\dbname-\type}
        \vspace{-5pt}
        \begin{subtable}{\linewidth}%
          \caption{$H_0: M_1 = M_2$, $H_a: M_1 \neq M_2$}%
          \label{app:tab:st-\dbname-\type-equal}
          \vspace{-5pt}
          \typesettable{./stadistics/tables/csv/\folder/two-sided-\dbname.csv}%
        \end{subtable}
        \begin{subtable}{\linewidth}%
          \vspace{5pt}
          \caption{$H_0: M_1 \leq M_2$, $H_a: M_1 > M_2$}%
          \label{app:tab:st-\dbname-\type-greater}
          \vspace{-5pt}
          \typesettable{./stadistics/tables/csv/\folder/greater-\dbname.csv}%
        \end{subtable}%
      \end{table}%
    \end{landscape}%
}}

\pgfplotstableset{
  multistyler/.style 2 args={
    @my multistyler/.style={display columns/##1/.append style={#2}},
    @my multistyler/.list={#1}
  },
  model joined table/.style={
    col sep=comma,
    empty cells with={--}, %
    every column/.style={
      fixed, zerofill,
      precision=2,
    },
    multistyler={2,...,25}{%
      postproc cell content/.append code={%
        \ifstrequal{########1}{}{}{%
          \pgfmathparse{int(less(########1, 0.05))}%
          \ifnum\pgfmathresult=1%
          \pgfkeysalso{/pgfplots/table/@cell content/.add={\cellcolor{lightgray}}{}}%
          \fi%
        }%
      },
    },
    columns/db_name/.style={
      string type,
      column type=l,
      string replace={CK}{CK+},
      string replace={OULU}{OULU-Casia},
      assign cell content/.code={
        \pgfmathparse{int(Mod(\pgfplotstablerow,6))}%
        \ifnum\pgfmathresult=0%
        \pgfkeyssetvalue{/pgfplots/table/@cell content}{####1}%
        \else%
        \pgfkeyssetvalue{/pgfplots/table/@cell content}{ }%
        \fi%
      },
    },
    columns/init/.style={
      string type,
      column type=l,
      string replace={0000}{RI},
      string replace={0001}{RI+SD},
      string replace={0010}{RI+DA},
      string replace={1000}{FT},
      string replace={1001}{FT+SD},
      string replace={1010}{FT+DA},
    },
    every head row/.style={
      output empty row,
      before row={
        \toprule%
        && \multicolumn{6}{l}{CK+} & \multicolumn{6}{l}{MMI} & \multicolumn{6}{l}{OULU-Casia} & \multicolumn{6}{l}{MUG}\\%
        \cmidrule{3-26}%
        && \footnotesize RI & \footnotesize RI+DA & \footnotesize RI+SD & \footnotesize FT & \footnotesize FT+DA & \footnotesize FT+SD 
        & \footnotesize RI & \footnotesize RI+DA & \footnotesize RI+SD & \footnotesize FT & \footnotesize FT+DA & \footnotesize FT+SD 
        & \footnotesize RI & \footnotesize RI+DA & \footnotesize RI+SD & \footnotesize FT & \footnotesize FT+DA & \footnotesize FT+SD 
        & \footnotesize RI & \footnotesize RI+DA & \footnotesize RI+SD & \footnotesize FT & \footnotesize FT+DA & \footnotesize FT+SD\\%
      },
      after row=\midrule,
    },
    every last row/.style={after row=\bottomrule},
    every even row/.style={
      before row={\rowcolor[gray]{0.95}}
    },
  },
}
\newcommand{\typesettablemodeljoined}[1]{%
  \resizebox{\linewidth}{!}{%
    \pgfplotstabletypeset[%
    model joined table,
    ]{#1}}
}

\foreach \folder/\type in {auto_ablation_merge_models/single-models-unified,cross_kfold_ablation_merge_models/merged-models-unified,mix-merge_models/single-vs-merged-models-unified}{%
  \begin{landscape}%
    \begin{table}%
      \caption{$p$-values of the hypothesis testing for (a)~$H_0: M_1 = M_2$, $H_a: M_1 \neq M_2$ and (b)~$H_0: M_1 \leq M_2$, $H_a: M_1 > M_2$ for \type-database accuracies, where $M_1$ are the methods on the left and $M_2$ are the methods on top.  The shaded cells denote the rejection of the null hypothesis at $5\%$ significance level.}%
      \label{app:tab:st-\type}
      \vspace{-5pt}
      \begin{subtable}{\linewidth}%
        \caption{$H_0: M_1 = M_2$, $H_a: M_1 \neq M_2$}%
        \label{app:tab:st-\type-equal}
        \vspace{-5pt}
        \typesettablemodeljoined{./stadistics/tables/csv/\folder/joined-two-sided.csv}%
      \end{subtable}
      \begin{subtable}{\linewidth}%
        \vspace{5pt}
        \caption{$H_0: M_1 \leq M_2$, $H_a: M_1 > M_2$}%
        \label{app:tab:st-\type-greater}
        \vspace{-5pt}
        \typesettablemodeljoined{./stadistics/tables/csv/\folder/joined-greater.csv}%
      \end{subtable}%
    \end{table}%
  \end{landscape}%
}

\pgfplotstableset{
multistyler/.style 2 args={
  @my multistyler/.style={display columns/##1/.append style={#2}},
  @my multistyler/.list={#1}
},
hypothesis_merge_model_and_dataset table/.style={
  col sep=comma,
  empty cells with={--}, %
  every column/.style={
    fixed, zerofill,
    precision=2,
  },
	multistyler={1,...,6}{%
		postproc cell content/.append code={%
			\ifstrequal{########1}{}{}{%
				\pgfmathparse{int(less(########1, 0.05))}%
				\ifnum\pgfmathresult=1%
				\pgfkeysalso{/pgfplots/table/@cell content/.add={\cellcolor{lightgray}}{}}%
				\fi%
			}%
		},
	},
	columns/init/.style={
		string type,
		column type=l,
		string replace={0000}{RI},
		string replace={0001}{RI+SD},
		string replace={0010}{RI+DA},
		string replace={1000}{FT},
		string replace={1001}{FT+SD},
		string replace={1010}{FT+DA},
	},
  every head row/.style={
    output empty row,
    before row={
      \toprule%
      & \footnotesize RI & \footnotesize RI+DA & \footnotesize RI+SD & \footnotesize FT & \footnotesize FT+DA & \footnotesize FT+SD \\%
    },
    after row=\midrule,
  },
  every last row/.style={after row=\bottomrule},
  every even row/.style={
    before row={\rowcolor[gray]{0.95}}
  },
},
}
\newcommand{\typesettablemodelanddatasetmerge}[1]{%
\resizebox{
\ifdim\width>\linewidth%
  \linewidth%
\else%
  \width%
\fi%
}{!}{%
\pgfplotstabletypeset[%
  hypothesis_merge_model_and_dataset table,
]{#1}}
}
  \foreach \folder/\type in {auto_ablation_merge_models_and_datasets/single-models unified-dataset,cross_kfold_ablation_merge_models_and_datasets/merged-models unified-dataset,mix-merge_models_and_datasets/single-vs-merged unified-models and -dataset}{%
    \begin{landscape}%
      \begin{table}%
        \caption{$p$-values of the hypothesis testing for (a)~$H_0: M_1 = M_2$, $H_a: M_1 \neq M_2$ and (b)~$H_0: M_1 \leq M_2$, $H_a: M_1 > M_2$ on \type\ accuracies, where $M_1$ are the methods on the left and $M_2$ are the methods on top.  The shaded cells denote the rejection of the null hypothesis at $5\%$ significance level.}%
        \label{app:tab:st-merged-\type}
        \vspace{-5pt}
        \begin{subtable}{\linewidth}%
          \centering%
          \caption{$H_0: M_1 = M_2$, $H_a: M_1 \neq M_2$}%
          \label{app:tab:st-merged-\type-equal}
          \vspace{-5pt}
          \typesettablemodelanddatasetmerge{./stadistics/tables/csv/\folder/two-sided.csv}%
        \end{subtable}
        \begin{subtable}{\linewidth}%
          \centering%
          \vspace{5pt}
          \caption{$H_0: M_1 \leq M_2$, $H_a: M_1 > M_2$}%
          \label{app:tab:st-merged-\type-greater}
          \vspace{-5pt}
          \typesettablemodelanddatasetmerge{./stadistics/tables/csv/\folder/greater.csv}%
        \end{subtable}%
      \end{table}%
    \end{landscape}%
}

\end{document}